\documentclass{article}

    \usepackage[preprint,nonatbib]{neurips_2022}

\usepackage[utf8]{inputenc} 
\usepackage[T1]{fontenc}    
\usepackage{hyperref}       
\usepackage{url}            
\usepackage{booktabs}       
\usepackage{amsfonts}       
\usepackage{nicefrac}       
\usepackage{microtype}      
\usepackage{xcolor}         

\usepackage{amsmath}
\usepackage{graphicx}
\usepackage{rotating}
\usepackage{tabulary}
\usepackage{algorithm}
\usepackage{algorithmic}
\usepackage{subdef}
\usepackage{wrapfig,lipsum,booktabs}
\usepackage{subcaption}

\usepackage{multirow}
\usepackage{longtable}

\newcommand{\figref}[1]{Fig.~\ref{#1}}
\newcommand{\tabref}[1]{Tab.~\ref{#1}}
\newcommand{\secref}[1]{Sec.~\ref{#1}}
\newcommand{\AlgRef}[1]{Algo.~\ref{#1}}
\newcommand{\equref}[1]{Equ.~(\ref{#1})}
\newcommand{\Appref}[1]{Appendix.~\ref{#1}}
\newcommand{\model}{\mbox{\textsc{Add}}}

\newcommand{\smi}{\mbox{\textsc{Smi}}}
\newcommand{\scg}{\mbox{\textsc{Scg}}}
\newcommand{\scmi}{\mbox{\textsc{Scmi}}}

\title{Active Data Discovery: Mining Unknown Data using Submodular Information Measures}

\author{Suraj Kothawade \\ University of Texas at Dallas \\ \texttt{suraj.kothawade@utdallas.edu}
\And
Shivang Chopra \\ Microsoft Research \\ \texttt{t-schopra@microsoft.com}
\And
Saikat Ghosh \\ University of Texas at Dallas \\ \texttt{saikat.ghosh@utdallas.edu}
\And
Rishabh Iyer \\ University of Texas at Dallas \\ \texttt{rishabh.iyer@utdallas.edu}
}

\begin{document}

\maketitle

\begin{abstract}
Active Learning is a very common yet powerful framework for iteratively and adaptively sampling subsets of the unlabeled sets with a human in the loop with the goal of achieving labeling efficiency. Most real world datasets have imbalance either in classes and slices, and correspondingly, parts of the dataset are rare. As a result, there has been a lot of work in designing active learning approaches for mining these rare data instances. Most approaches assume access to a seed set of instances which contain these rare data instances. However, in the event of more extreme rareness, it is reasonable to assume that these rare data instances (either classes or slices) may not even be present in the seed labeled set, and a critical need for the active learning paradigm is to efficiently discover these rare data instances. In this work, we provide an active data discovery framework which can mine unknown data slices and classes efficiently using the submodular conditional gain and submodular mutual information functions. We provide a general algorithmic framework which works in a number of scenarios including image classification and object detection and works with both rare classes and rare slices present in the unlabeled set. We show significant accuracy and labeling efficiency gains for unknown classes ($\approx 10\% - 15\%$) and unknown slices ($\approx 5\% - 7\%$) with our approach compared to existing state-of-the-art active learning approaches for actively discovering these unknown classes and slices.  
\end{abstract}

\section{Introduction}

\begin{wrapfigure}{R}{0.4\textwidth}
\includegraphics[width=0.4\textwidth]{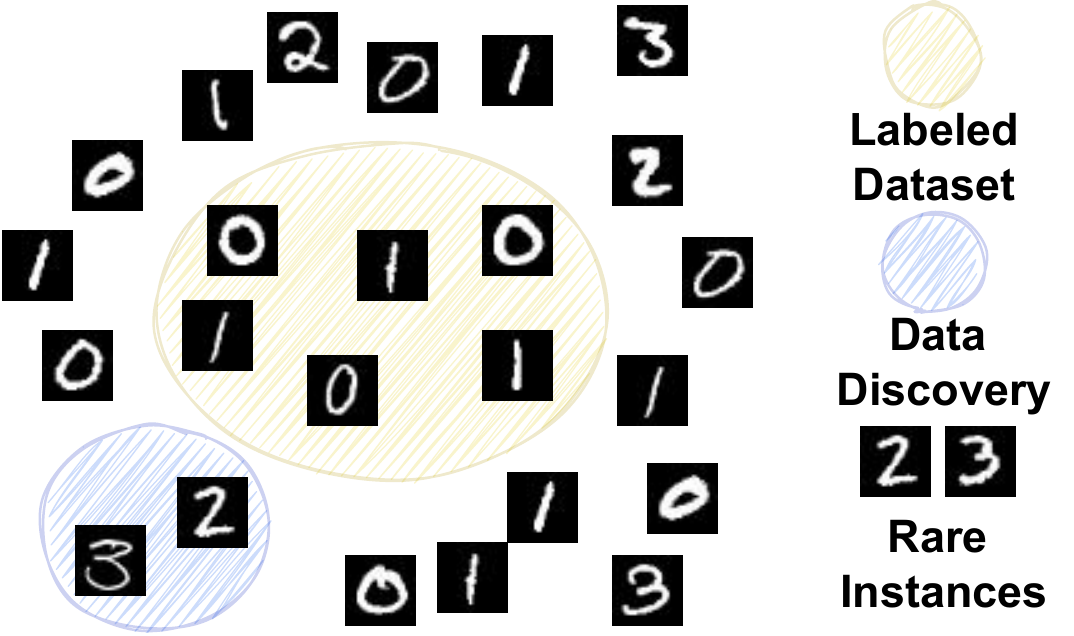}
\caption{\textbf{The data discovery problem.} Digits 0 and 1 are known in the labeled dataset. Unlabeled 
dataset has additional digits 2 and 3 which are rare instances and unknown to the labeled dataset. 
\vspace{-2.5ex}
}
\label{fig:intro}
\end{wrapfigure}

Machine learning based predictions have been widely used in critical real-world domains like medical imaging and autonomous driving. Their success relies on availability of suitable supervised training data that can represent potential scenarios during test time. Unfortunately, real-world datasets are imbalanced and contain rare instances of data. These rare instances could represent a class (\eg\ cat) or a data slice (\eg\ brown animals). Models that are trained by using these imbalanced datasets are biased and perform poorly on these rare instances. A common practice to mitigate this imbalance is to iteratively acquire more \textit{labeled} data by sampling from an \textit{unlabeled} dataset by using human-in-the-loop active learning (AL) strategies. However, these techniques require a few exemplars of rare instances and assume knowledge about the total number of classes. In this paper, we study a new scenario of extreme imbalance such that the rare instances are completely absent from the labeled training dataset. Furthermore, we do not know if they even exist in the unlabeled dataset. Hence, we call such instances as \emph{unknown} instances and the problem of finding them as the \emph{data discovery} problem. This problem addresses the following question: \emph{Can data points of unknown instances be discovered from a large unlabeled dataset in order to train a robust machine learning model?}

In \figref{fig:intro}, we present an illustration of the data discovery problem. Initially, the labeled dataset has digits 0 and 1, whereas the larger unlabeled dataset comprises digits from 0-3. The digits 2 and 3 are absent in the initial labeled set, as they are rare instances. Similar phenomena occur in many real-world domains. For instance, in medical imaging certain cancers are much rarer (few tens) than others (few thousands) \cite{codella2019skin, aptos2019}. Another example is in the autonomous driving domain, where certain objects and slices might be extremely rare (\eg traffic signs in foggy weather). Naturally, some of these rare instances are likely to be missing in the labeled training dataset. As a result, these rare instances will be \emph{unknown} to a model that is trained using such a dataset, and they will \emph{always} be misclassified during test time. As a solution to this problem, we propose \model, an active learning based data discovery framework that iteratively finds data points of unknown instances, which can then be labeled by a human-in-the-loop and added to the labeled training dataset.

\subsection{Related work} \label{sec:related_work}

\textbf{Active Learning (AL).} Active learning is a widely used approach to select informative data points from an unlabeled dataset to be labeled in an iterative manner with a human-in-the-loop. Uncertainty based AL methods aim to select the most uncertain data points for labeling. Intuitively, these uncertainty based methods should ideally select data points of unknown instances, but as we shall see in \secref{sec:experiments}, they also select data points from known instances that the model isn't confident about. The most common uncertainty based AL technique is \textsc{Entropy}~\cite{settles2009active} that aims to select data points with maximum entropy. Other techniques that are widely used are least-confidence (\textsc{Least-Conf}) \cite{wang2014new}, which selects points with lowest-confidence, and \textsc{Margin} \cite{roth2006margin}, which selects points that have the lowest difference in confidence of the top two predictions. The main drawback of uncertainty based methods is that they lack diversity within the acquired subset. To mitigate this, a number of approaches have proposed to incorporate diversity. A recent approach called \textsc{Badge}~\cite{ash2020deep} uses the last linear layer gradients computed using pseudo-labels to represent data points and runs \textsc{K-means++}~\cite{kmeansplus} to obtain centers that have a high gradient magnitude. The centers being representative and having high gradient magnitude ensures uncertainty and diversity at the same time. However, for data discovery, computing these pseudo-labels fairly is \emph{not} possible since the unknown instances will always be incorrectly pseudo-labeled. 
Another method, \textsc{BatchBald}~\cite{kirsch2019batchbald} requires a large number of Monte Carlo dropout samples to obtain significant mutual information which limits its application to data discovery where data points from unknown classes are unavailable.
Recently, \cite{kothawade2021similar} proposed the use of submodular information measures for active learning in realistic scenarios, while~\cite{kothawade2021talisman} used them to find rare objects in an autonomous driving object detection problem. Their method focuses on acquiring data points from the rare instances and assumes a set of data points that represent \emph{all} the rare instances. Our proposed method, \model, does not assume any knowledge about the rare/unknown instances, and hence, it can be broadly applied for data discovery. \model\ uses a combination of submodular conditional gain (\scg) and submodular mutual information (\smi) functions to iteratively find data points of unknown instances. We define these functions in \secref{sec:preliminaries} and present our framework in \secref{sec:our_method}.

\textbf{Exploratory Machine Learning.}
There exists some work in the exploratory machine learning domain that is slightly related to the data discovery problem. The differences in the problem settings are either in the \textit{definition} of unknown instances or the \textit{underlying reasons} that make these instances are unknown. \cite{attenberg2015beat, lakkaraju2017identifying} define unknown instances as data points that are misclassified with high confidence. Their problem setting is different in that they assume that all possible classes are known and high-confidence false-predictions are unknowns, whereas the data discovery problem assumes no knowledge about the total number of classes and defines unknown instances to be \emph{all} data points from unknown classes. In another work, \cite{zhao2021exploratory} propose a two-step process, where they consider unknown instances to be the data points that are classified with low-confidence. They propose modifying the feature space by adding \emph{new features} to learn a better decision boundary for classifying the low-confidence data points. In contrast, our active data discovery framework aims at adding \emph{new data points} by sampling them from a large unlabeled dataset without modifying the original feature space. Moreover, our approach considers all data points in the unlabeled dataset and is not limited to considering only low-confidence data points. This is critical, since unknown instances can also be misclassified with high-confidence.

\subsection{Our contributions} \label{sec:contributions}
We summarize our contributions as follows: \textbf{1)} We introduce the problem of data discovery where unknown instances are absent in the labeled training dataset that leads to failure on critical deployment cases. \textbf{2)} Given the limitations of active learning and exploratory machine learning, we propose \model, a novel active data discovery framework that can be used to effectively discover unknown instances from a large pool of unlabeled data. \textbf{3)} We demonstrate the effectiveness of our framework for unknown classes and slices in the context of image classification and for unknown object classes in the context of object detection. Specifically, we show that \model\ discovers the most number of unknown instances and obtains $\approx 10\% - 15\%$ gain in average accuracy of unknown classes and $\approx 5\% - 7\%$ gain in average accuracy of unknown slices over the best performing baseline for image classification (see \secref{sec:exp_ic}). For object detection, \model\ obtains $\approx 5\% - 7\%$ gain in average AP of unknown classes over the best performing baseline (see \secref{sec:exp_od}).

\section{Preliminaries} \label{sec:preliminaries}

In this section, we discuss the submodular functions that are used as acquisition functions in the proposed active data discovery framework. 

\textbf{Submodular Functions: } We let $\Vcal$ denote the \emph{ground-set} of $n$ data points $\Vcal = \{1, 2, 3,...,n \}$ and a set function $f:
 2^{\Vcal} \xrightarrow{} \mathbb{R}$.  
 These functions can be optimized in \textit{near-linear time} using a greedy algorithm \cite{mirzasoleiman2015lazier}. Facility location, graph cut, log determinants, {\em etc.} are some examples \cite{iyer2015submodular}. 

\begin{table*}[!htb]
 
   
    \begin{subtable}{.45\linewidth}
      \centering
        \caption{Instantiations of \smi\ functions}
        \label{tab:smi_inst}
        \begin{tabular}{|c|c|c|}
        \hline
        \textbf{\smi} & \textbf{$I_f(\Acal;\Qcal)$} \\ \hline
        \scriptsize{\textsc{Flmi}}             & \scriptsize{$\sum\limits_{i \in \Qcal} \max\limits_{j \in \Acal} S_{ij} + $ $ \eta \sum\limits_{i \in \Acal} \max\limits_{j \in \Qcal} S_{ij}$}                \\
        \scriptsize{\textsc{Gcmi}}              & \scriptsize{$2 \lambda \sum\limits_{i \in \Acal} \sum\limits_{j \in \Qcal} S_{ij}$}                \\
        \scriptsize{\textsc{Logdetmi}}          & \scriptsize{$\log\det(S_{\Acal}) -\log\det(S_{\Acal} -$} \\ & \scriptsize{$ \eta^2 S_{\Acal,\Qcal}S_{\Qcal}^{-1}S_{\Acal,\Qcal}^T)$}          
        \\ \hline              
        \end{tabular}
    \end{subtable}%
    \begin{subtable}{.6\linewidth}
      \centering
        \caption{Instantiation of \scg\ and \scmi\ functions}
        \label{tab:scg_scmi_inst}
        \begin{tabular}{|c|c|}
        \hline
        \textbf{\scg} & \textbf{$f(\Acal|\Pcal)$} \\ \hline
        \scriptsize{\textsc{Flcg}}       & \scriptsize{$\sum\limits_{i \in \Vcal} \max(\max\limits_{j \in \Acal} S_{ij}-$ $ \max\limits_{j \in \Pcal} S_{ij}, 0)$}                \\
        \scriptsize{\textsc{Logdetcg}}       & \scriptsize{$\log\det(S_{\Acal} - \nu^2 S_{\Acal, \Pcal}S_{\Pcal}^{-1}S_{\Acal, \Pcal}^T)$} \\
        \scriptsize{\textsc{Gccg}}   & \scriptsize{$f(\Acal) - 2 \lambda \nu \sum\limits_{i \in \Acal, j \in \Pcal} S_{ij}$}                \\ \hline
        \end{tabular}
        \vspace{0.5ex}
        
        \begin{tabular}{|c|c|}
        \hline
        \textbf{\scmi} & \textbf{$I_f(\Acal;\Qcal|\Pcal)$} \\ \hline
        \scriptsize{\textsc{Flcmi}}       & \scriptsize{$\sum\limits_{i \in \Vcal} \max(\min(\max\limits_{j \in \Acal} S_{ij},$ $ \max\limits_{j \in \Qcal} S_{ij})$} \scriptsize{$-  \max\limits_{j \in \Pcal} S_{ij}, 0)$}                \\\scriptsize{\textsc{Logdetcmi}}   & \scriptsize{$\log \frac{\det(I - S_{\Pcal}^{-1}S_{\Pcal, \Qcal} S_{\Qcal}^{-1}S_{\Pcal, \Qcal}^T)}{\det(I - S_{\Acal \cup \Pcal}^{-1} S_{\Acal \cup \Pcal, Q} S_{\Qcal}^{-1} S_{\Acal \cup \Pcal, Q}^T)}$} \\ \hline               
        \end{tabular}
    \end{subtable} 
    \caption{Instantiations of different \scg, \smi, \scmi\ functions that can be used as acquisition functions for data discovery. }
    \label{tab:SIM_inst}
\end{table*}

\noindent \textbf{Submodular Conditional Gain (\scg).} Given sets $\Acal, \Pcal \subseteq \Ucal$, the \scg\ $f(\Acal | \Pcal)$ is the gain in function value of $f$ by adding $\Acal$ to $\Pcal$. Thus, $f(\Acal | \Pcal) = f(\Acal \cup \Pcal) - f(\Pcal)$ \cite{iyer2021submodular}. Intuitively, \scg\ models how different $\Acal$ is from $\Pcal$, and maximizing \scg\ functions will select data points \emph{dissimilar} to the points in $\Pcal$ while being diverse. The parameter $\nu$ is used to control the intensity of conditioning of the selected subset $\Acal$ with $\Pcal$. We refer to $\Pcal$ as the conditioning set. We use the \scg\ functions in the \model\ framework for discovering unknown instances by conditioning on the known instances.

\noindent \textbf{Submodular Mutual Information (\smi):} Given a set of items $\Acal, \Qcal \subseteq \Vcal$, the \smi\ \cite{levin2020online,iyer2021submodular, kothawade2021prism} is defined as $I_f(\Acal; \Qcal) = f(\Acal) + f(\Qcal) - f(\Acal \cup \Qcal)$. Intuitively, this measures the similarity between $\Qcal$ and $\Acal$, and we refer to $\Qcal$ as the query set. The parameter $\eta$ is used to control the intensity of relevance of the selected subset $\Acal$ with the query set $\Qcal$. In the \model\ framework, the \smi\ functions are used to find more data points of unknown instances that are semantically similar to a set of unknown instances found by \scg. 

\noindent \textbf{Submodular Conditional Mutual Information (\scmi):} 
Given sets $\Acal, \Qcal, \Pcal \subseteq \Ucal$, the \scmi\ is defined as $I_f(\Acal; \Qcal | \Pcal) = f(\Acal \cup \Pcal) + f(\Qcal \cup \Pcal) - f(\Acal \cup \Qcal \cup \Pcal) - f(\Pcal)$. Intuitively, \scmi\ jointly models the similarity between $\Acal$ and $\Qcal$ and their dissimilarity, with $\Pcal$. Similar to \smi\ and, \scg\ this can be tuned by using parameters $\eta$ and $\nu$ respectively. We do not add them in the \scmi\ equations for simplicity.

\noindent \textbf{Instantiations of different submodular functions for active data discovery:}
In \tabref{tab:SIM_inst}, we present the different submodular functions that are used for active data discovery. The formulations for facility location (\textsc{Fl}), graph cut (\textsc{Gc}) and log determinant (\textsc{Logdet}) are as in \cite{iyer2021submodular, kothawade2021prism}, and we adapt them as AL based acquisition functions for data discovery. Each function in \tabref{tab:SIM_inst} is named as the underlying submodular function, followed by \textsc{mi}/\textsc{cg}/\textsc{cmi}. For instance, the \textsc{Fl} based \smi\ function is denoted as \textsc{Flmi}, the \scg\ function is denoted as \textsc{Flcg}, and \scmi\ as \textsc{Flcmi}. The \textsc{Gc} and \textsc{Logdet} functions are denoted similarly. These functions are instantiated using a pairwise similarity matrix $S$, where $S_{\Acal,\Bcal}$ denotes similarity between items in $\Acal$ and $\Bcal$, while $S_{ij}$ denotes the entry $(i,j)$ in $S$.

\section{\model: Our framework for Active Data Discovery} \label{sec:our_method}

In this section, we propose \model, a novel active learning based framework for data discovery. We show how \model\ uses a combination of conditioning via \scg\ functions and mutual information via \smi\ functions to effectively acquire data points of unknown instances. 

The main idea behind the data discovery framework follows a \emph{conditioning and targeting} strategy. Assuming that the unlabeled set $\Ucal$ contains some unknown instances, we find them by conditioning on $\Pcal$ that contains data points of only the known instances. Note that $\Pcal$ is initialized as $\Lcal$, since the initial labeled set has only known instances. Intuitively, by conditioning on $\Pcal$, we are trying to find data points that are \emph{dissimilar} to the known instances, thereby potentially finding unknown instances. We perform conditioning by maximizing the \scg\ function using a greedy algorithm \cite{mirzasoleiman2015lazier}:

\begin{align}\label{eq:SCG-al}
\max_{\Acal \subseteq \Ucal, |\Acal| \leq B} f(\Acal | \Pcal)    
\end{align}

The \scg\ function is instantiated using a similarity kernel $\Scal^\Pcal \in \mathbb{R}^{|\Pcal| \times |\Ucal|}$ that contains pairwise similarities between data points in $\Pcal$ and $\Ucal$ in the feature space. In every round of AL, we obtain labels via a human-in-the-loop for the selected subset $\Acal$ and add it to the labeled set $\Lcal$. We augment the conditioning set $\Pcal$ with $\Acal^\Pcal \subseteq \Acal$ containing newly selected known instances, and $\Qcal$ is augmented with $\Acal^\Qcal \subseteq \Acal$ containing newly selected unknown instances. Note that $\Acal = \Acal^\Pcal \cup \Acal^\Qcal$. 

We keep a track of the unique known instances throughout AL rounds using a concept coverage set $\Kcal$. Typically, $\Kcal$ contains unique concepts like class indices if the goal is to discover classes, or $\Kcal$ may contain attributes (\eg\ color, shape), if they goal is to discover slices, or a combination of class and attribute (\eg\ brown cat). If there are no new concepts in the newly discovered subset $\Acal^C$, \ie\ $\Acal^C \cap \Kcal == \emptyset$, we conclude the conditioning phase and start the targeting phase.

\begin{algorithm}
\begin{algorithmic}[1]
\REQUIRE Initial Labeled set of data points: $\Lcal$, containing $\Kcal$ unique known instances. Large unlabeled dataset: $\Ucal$. Initial conditioning set $\Pcal \leftarrow \Lcal$, query set $\Qcal \leftarrow \emptyset$. Model $\Mcal$, batch size: $B$, number of selection rounds: $N$, $unknown = $ True \\
\FOR{selection round $i = 1:N$}
\STATE Train model $\Mcal$ with loss $\Hcal$ on the current labeled set $\Lcal$ and obtain parameters $\theta$
\IF{$unknown == $ True} 
    \STATE Compute $S^\Pcal \in \mathbb{R}^{|\Pcal| \times |\Ucal|}$ such that: $S_{pu} \leftarrow$ \textsc{Cosine\_Sim}($\Mcal_{\theta_i}, p, u$), $\forall p \in \Pcal, \forall u \in \Ucal$ 
    \STATE Instantiate a \scg\ function $f(\Acal|\Pcal)$ based on $S^\Pcal$.
    \STATE $\Acal_i \leftarrow \mbox{argmax}_{\Acal \subseteq \Ucal, |\Acal | \leq B}  f(\Acal| \Pcal)$  \textcolor{blue}{\{Maximize \scg\ if unknown instances may exist\}}

\ELSE 
    \STATE Compute $S^\Qcal \in \mathbb{R}^{|\Qcal| \times |\Ucal|}$ such that: $S_{qu} \leftarrow$ \textsc{Cosine\_Sim}($\Mcal_{\theta_i}, q, u$), $\forall q \in \Qcal, \forall u \in \Ucal$ 
    \STATE Instantiate a \smi\ function $I_f(\Acal;\Qcal)$ based on $S^\Qcal$.
    \STATE $\Acal_i \leftarrow \mbox{argmax}_{\Acal \subseteq \Ucal, |\Acal | \leq B}  I_f(\Acal; \Qcal)$  \textcolor{blue}{\{Maximize \smi\ if no more unknown instances exist can be found\}}
\ENDIF
\STATE Get labels $L(\Acal_i)$ for batch $\Acal_i$ and $\Lcal \leftarrow \Lcal \cup L(\Acal_i)$, $\Ucal \leftarrow \Ucal - \Acal_i$
\STATE $\Pcal \leftarrow \Pcal \cup \Acal_i^\Pcal, \Qcal \leftarrow \Qcal \cup \Acal_i^\Qcal$ \textcolor{blue}{\{Add newly selected known instances to $\Pcal$ and unknown instances to $\Qcal$\}}
\IF{$\Acal_i^C \cap \Kcal == \emptyset$}
\STATE $unknown = $ False \textcolor{blue}{\{No new unknown instances discovered\}}
\ENDIF
\STATE $\Kcal \leftarrow \Kcal \cup \Acal_i^C$ \textcolor{blue}{\{Add newly discovered unique instances, if any\}}
\ENDFOR
\STATE \textbf{Return} trained model $\Mcal_{\theta_N}$ and labeled set $\Lcal$ augmented with newly discovered instances.
\end{algorithmic}
\caption{\model: Active Data Discovery}
\label{algo:add}
\end{algorithm}

In the targeting phase, we use a query set $\Qcal$ containing unknown instances that were accumulated in the conditioning phase. We maximize the mutual information with $\Qcal$ to find more semantically similar unknown instances from the unlabeled set $\Ucal$. We do so by maximizing the \smi\ function using a greedy algorithm \cite{mirzasoleiman2015lazier}: 

\begin{align}\label{eq:SMI-al}
\max_{\Acal \subseteq \Ucal, |\Acal| \leq B} I_f(\Acal; \Qcal)    
\end{align}

Similar to the targeting phase, we augment $\Lcal, \Pcal$ and $\Qcal$ in every round of AL. Note that the purpose of the conditioning phase is to find a few points that represent the unknown instances and can serve as exemplars for the targeting phase. Note that one can continue conditioning throughout all AL rounds using, \scg\ (\equref{eq:SCG-al})  or simultaneously perform conditioning and targeting using \scmi\ (\equref{eq:SCMI-al}):

\begin{align}\label{eq:SCMI-al}
\max_{\Acal \subseteq \Ucal, |\Acal| \leq B} I_f(\Acal; \Qcal | \Pcal).    
\end{align}
 However, we empirically find the combination of \scg\ for conditioning and \smi\ for targeting to be most effective (see \secref{sec:experiments}) and scalable. We present our active data discovery framework in \AlgRef{algo:add}. The \model\ framework is generic and can be applied for any task. The only step that changes across tasks is the computation of pairwise similarity kernels $\Scal^\Pcal$ and $\Scal^\Qcal$. In this paper, we apply \model\ for image classification and object detection. Next, we discuss the similarity kernel computation for these tasks. \looseness-1

\textbf{Similarity kernel for image classification: }
In order to represent each data point, we extract features from the penultimate layer of the model $\Mcal$ that is trained using $\Lcal$. Next, the cosine similarity kernels for an image classification discovery task can be computed easily by computing the dot product between the feature vectors. One can efficiently use off-the-shelf functions\footnote{np.tensordot or torch.tensordot} for efficiently computing a vectorized dot product. 

\textbf{Similarity kernel for object detection:}
Computing the similarity kernel for data discovery in object detection tasks is slightly more complicated, since the similarities are computed at an object level. We borrow inspiration from \cite{kothawade2021talisman}, and compute object-level similarity by using the ground-truth bounding boxes for images in the labeled dataset and region proposals for images in the unlabeled dataset.

For simplicity, we shall describe the similarity computation for, $S_{ij}$ which represents the similarity between a single labeled image $i \in \Pcal$ (for $\Scal^\Pcal$) or $i \in \Qcal$ (for $\Scal^\Qcal$) and a single unlabeled image $j \in \Ucal$. Let $T$ be the number of ground-truth (GT) bounding boxes for $i$ and $R$ be the number of proposals obtained from a region-proposal-network (RPN) for $j$. Next, we extract features for $\Tcal$ GT boxes to obtain $\Fcal^{T \times D}$ and $\Rcal$ proposals to obtain $\Fcal^{R \times D}$, where $D$ is the dimensionality of the feature vector. These proposal features are extracted using the region-of-interest (RoI) pooling layer of the detection model $\Mcal$ trained on $\Lcal$. Next, we compute a dot-product between them along the feature dimension to obtain a GT-proposal score map $\Xcal \in \mathbb{R}^{\Tcal \times \Rcal}$.

Obtaining the final similarity value $S_{ij}$ from $\Xcal_{ij}$ differs for $S^\Pcal$ and $S^\Qcal$. For $S^\Qcal$, \ie\ if $i \in \Qcal$, we simply compute the element-wise maximum of $\Xcal_{ij}$. Intuitively, this $S^\Qcal_{ij}$ represents the best matching pair of GT box in $i$ and proposal in $j$. For $S^\Pcal$, \ie\ if $i \in \Pcal$, we first compute the maximum matching proposal for each GT box, followed by a min on the resulting vector. Intuitively, this $S^\Pcal_{ij}$ represents the worst matching GT-proposal pair, indicating that the proposal contains a potential unknown instance. We present similarity computation for a detection discovery task in \AlgRef{algo:detectionDiscoverySim}.

\begin{algorithm}
\begin{algorithmic}[1]
\REQUIRE Local feature extraction model $F_\theta$, $i \in \Qcal$ or $i \in \Pcal$ with $T$ GT boxes and $j \in \Ucal$ with $R$ region proposals. \\
\STATE $\Ecal_i \leftarrow $ $F_\theta(i$) \textcolor{blue}{\{$\Ecal_i \in \mathbb{R}^{T \times D}$\}}
\STATE $\Ecal_j \leftarrow $ $F_\theta(j$) \textcolor{blue}{\{$\Ecal_j \in \mathbb{R}^{R \times D}$\}}
\STATE $\Xcal_{ij} \leftarrow $ \textsc{Cosine\_Sim}($\Ecal_i, \Ecal_j$) \textcolor{blue}{\{$\Xcal_{ij} \in \mathbb{R}^{T \times R}$. Compute Cosine similarity along the feature dimension\}}
\IF {$i \in \Qcal$}
    \STATE $S_{ij} \leftarrow \max(\Xcal_{ij})$ \textcolor{blue}{\{Element-wise Max, $S_{ij}$ represents the score between the best matching pair of GT box $i \in \Qcal$, and query RoI $j \in \Ucal$\}} 
\ENDIF
\IF {$i \in \Pcal$}
    \STATE $S_{ij} \leftarrow \min(\max(\Xcal_{ij}))$ \textcolor{blue}{\{Max followed by Min, $S_{ij}$ represents the worst matching GT-proposal pair.\}}
\ENDIF
\STATE \textbf{Return} Similarity score $S_{ij}$ 
\end{algorithmic}
\caption{Detection Discovery Similarity}
\label{algo:detectionDiscoverySim}
\end{algorithm}

Note that $\Pcal$ contains GT boxes of objects that are known instances only, and $\Qcal$ contains GT boxes of objects from the unknown instances only. Since the similarities are computed at an object level, there might be an overlap in the query and conditioning set at an image level but \emph{not} at the object level.



\textbf{Scalability of \model: } We first note that \scg\ + \smi\ is more scalable compared to \scg\ + \scmi\ mainly because we do not need the additional compute due to $\Pcal$. Additionally, many \smi\ functions including \textsc{Flmi} do not require a $\Ucal \times \Ucal$ kernel but only require a $\Qcal \times \Ucal$ kernel ($\Qcal$ is often much smaller compared to $\Ucal$). Finally, we note that the \scg\ functions do require computing the $\Ucal \times \Ucal$ kernel. We can use the partitioning trick shown in~\cite{kothawade2021similar} where we basically partition the unlabeled set into partitions and operate on each partition individually (or even in parallel).

\section{Experiments} \label{sec:experiments}

In this section, we empirically evaluate the effectiveness of \model\ on a diverse set of datasets for image classification \secref{sec:exp_ic} and object detection (\secref{sec:exp_od}). Using these experiments, we show that existing AL approaches are not efficient for data discovery and that a \emph{conditioning and targeting} approach as proposed in \model\ is essential. In addition to standard bench-marking datasets, we also conduct experiments on a real-world medical dataset for discovering unknown classes (\secref{sec:u_classes}), and for a realistic unknown slices setting (\secref{sec:u_slices}). \secref{sec:exp_ic} In \Appref{app:add_exp}, we present an ablation study to show the effect of varying the parameters $\nu$ from \scg\ and $\eta$ from \smi\ that are used for controlling the degree of conditioning and targeting, respectively. For all experiments using the \model\ framework, we use the same underlying submodular function $f$ for \scg\ and \smi\ and is denoted as \textsc{Scg+mi} in the legend. For \eg, we use facility location based variants \textsc{Flcg} and \textsc{Flmi} for one experiment and denote it as \textsc{Flcg+mi}. Note that we use the same $f$ for simplicity of the experiments, and this is not a requirement for the \model\ framework. To ensure a fair treatment for all the methods, we use a common training procedure and set of hyper-parameters. We run all experiments $3 \times$ on a V100 GPU and provide error bars (standard deviation). 

\textbf{Baselines in all experiments:} 
We compare \model\ with several uncertainty and diversity based baselines since they are the most intuitive solutions for the data discovery problem. Particularly, we compare against three uncertainty based baselines: \textsc{Entropy} \cite{settles2009active}, \textsc{Margin} \cite{roth2006margin} and \textsc{Least-Conf} \cite{wang2014new}, and a recent diversity based baseline, \textsc{Badge} \cite{ash2020deep}. We discuss all these baselines in \secref{sec:related_work}. Lastly, we compare with \textsc{Random} sampling. 

\subsection{Image Classification} \label{sec:exp_ic}

In this section, we present the results for data discovery in the context of image classification tasks. We evaluate the performance of \model\ with existing AL acquisition functions for discovering unknown classes (\secref{sec:u_classes}) and unknown slices (\secref{sec:u_slices}). We do so by: 1) comparing the cumulative number of unknown data points selected by each acquisition function, and 2) comparing the mean accuracy obtained for the unknown classes or slices by training a model with the selected data points. For all the classification experiments, we train a ResNet-18 \cite{he2016deep} model using an SGD optimizer with an initial learning rate of 0.01, momentum of 0.9 and weight decay of 5e-4. In every round of AL, we reinitialize the weights of our model using Xavier initialization and train the model till 99\% accuracy is reached, or 200 epochs are complete. For evaluation, we use the default test set which contains both, the known and unknown instances.

\subsubsection{Unknown Classes} \label{sec:u_classes}

\noindent \textbf{Datasets and Experimental Setup:} For discovering unknown classes, we apply our framework to the standard MNIST \cite{lecun2010mnist} and CIFAR-10 \cite{krizhevsky2009learning} bench-marking datasets. We also conduct experiments on Path-MNIST \cite{medmnistv2, kather2019predicting}, a real-world medical imaging dataset for colorectal cancer classification. To consider a realistic scenario for data discovery, we create a labeled set $\Lcal$ containing data points from $K$ randomly chosen known classes. The unlabeled set $\Ucal$ contains data points from $X$ classes, where $X=K+Y$, \ie\ $\Ucal$ contains $Y$ additional unknown classes. Realistically, $\Ucal$ needs to be imbalanced and have lesser number of data points that belong to the unknown classes. Hence, we create an imbalance in the unlabeled set such that $|\Ucal_k| = \rho |\Ucal_y|$, where $k$ is a class from the $K$ known classes and $y$ is a class from the $Y$ unknown classes, and $\rho$ is an imbalance factor. For MNIST and CIFAR-10 ($X=10$), we set the first 7 classes as known ($K=7$) and the last three as unknown classes ($Y=3$). The number of data points in the labeled dataset $|\Lcal|=7000$, and unlabeled dataset $|\Ucal|=7150$ using an imbalance factor $\rho=20$, and a batch size $B=50$. For CIFAR10, we use $|\Lcal|=7000$, and $|\Ucal|=21900$ using an imbalance factor $\rho=10$ and $B=50$. For Path-MNIST, there exist a total of 9 classes ($X=9$), we set the first 7 classes as known ($K=7$) and the last two as unknown classes ($Y=2$). We use $|\Lcal|=3500$, and $|\Ucal|=7200$ and an imbalance factor $\rho=10$.

\begin{figure*}
\centering
\includegraphics[width = 14cm, height=1cm]{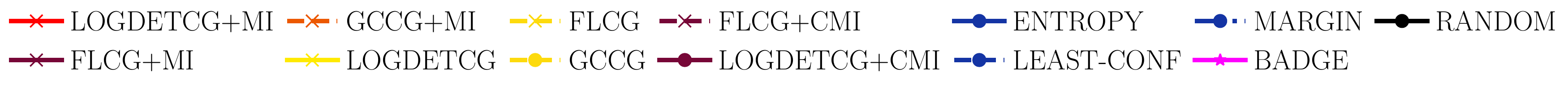}
\centering
\hspace*{-0.6cm}
\begin{subfigure}[t]{0.24\textwidth}
\includegraphics[width = \textwidth]{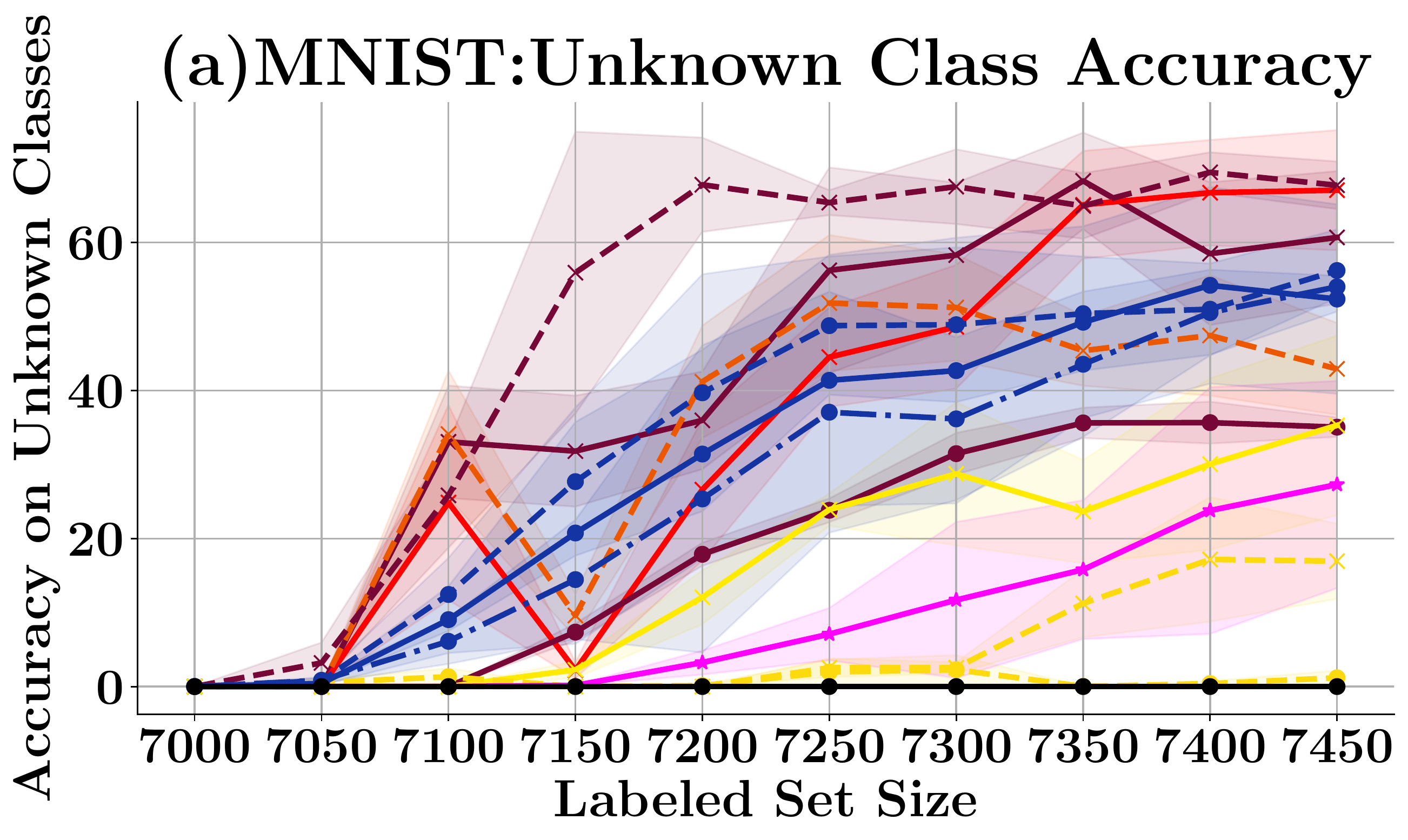}
\end{subfigure}
\begin{subfigure}[t]{0.25\textwidth}
\includegraphics[width = \textwidth]{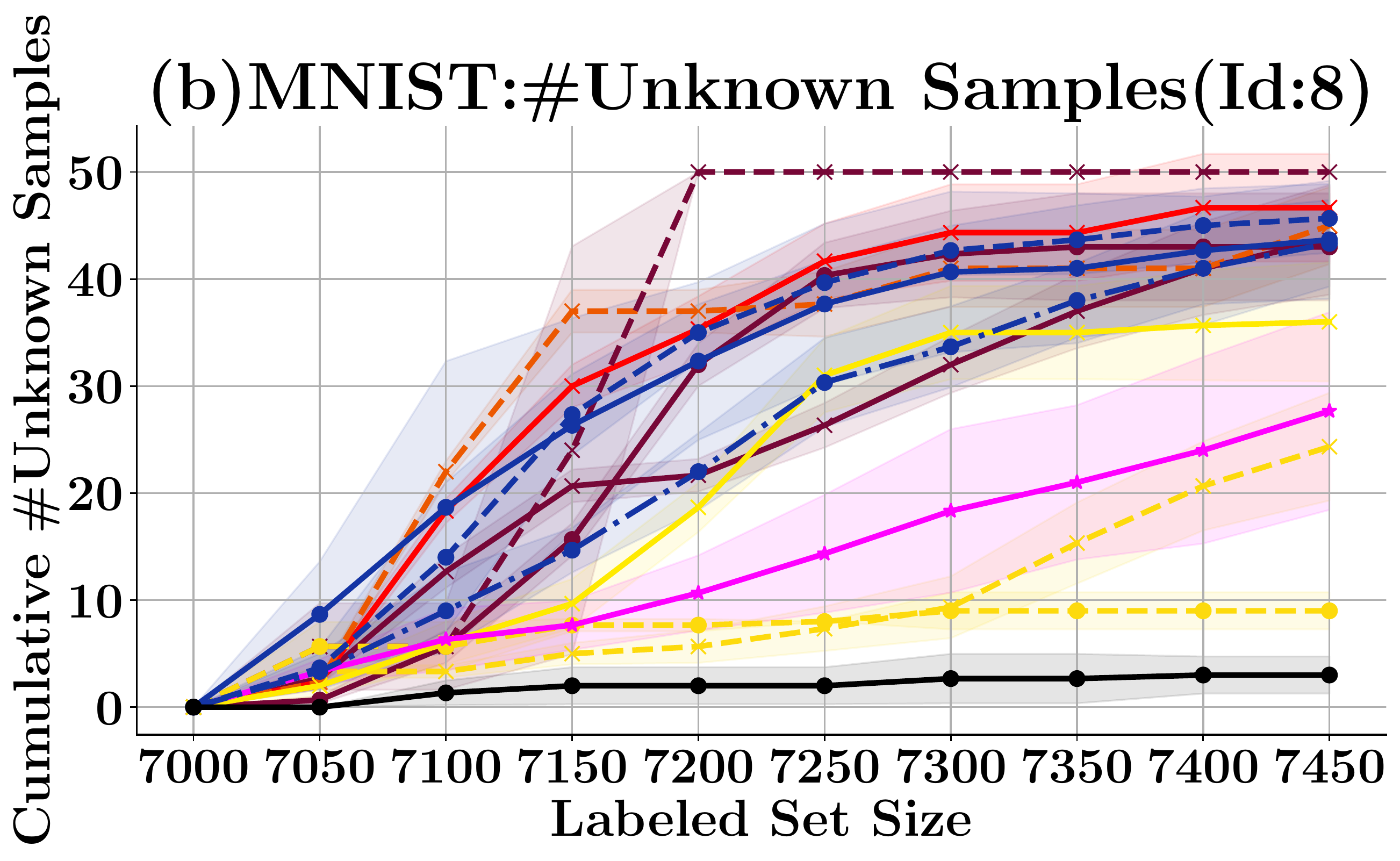}
\end{subfigure}
\begin{subfigure}[t]{0.25\textwidth}
\includegraphics[width = \textwidth]{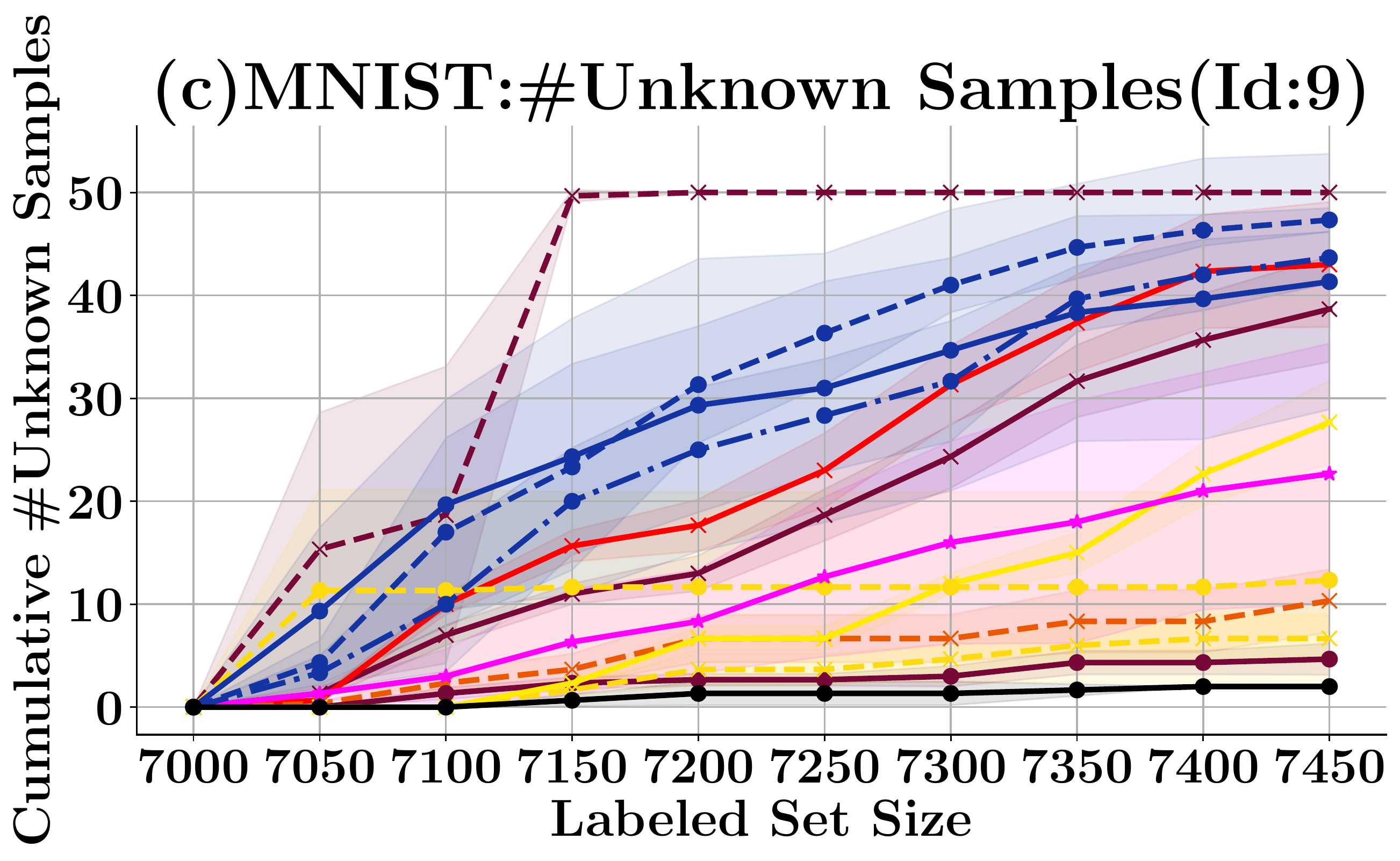}
\end{subfigure}
\begin{subfigure}[t]{0.25\textwidth}
\includegraphics[width = \textwidth]{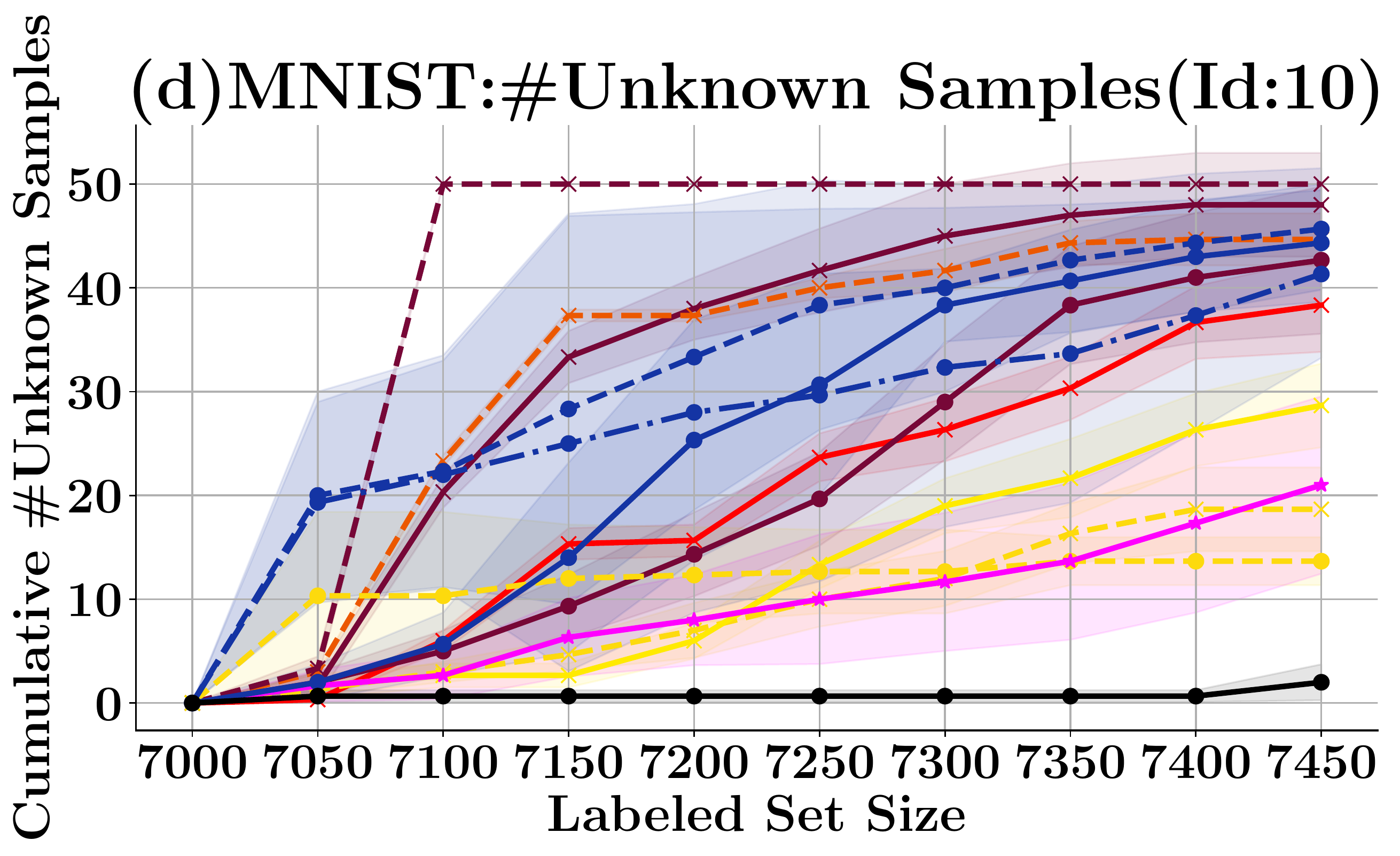}
\end{subfigure}
\caption{Active Data Discovery for unknown classes on MNIST. We observe that the \textsc{Scg+mi} (\textsc{Flcg+mi} and \textsc{Logdetcg+mi}) and \textsc{Scg+cmi} (\textsc{Flcg+cmi}) variants outperform other methods in terms of the average accuracy on the unknown classes. \textsc{Flcg+cmi} selects all data points from the unknown classes, the fastest, by $5^{th}$ round of AL.
\vspace{-1ex}
}
\label{fig:res_mnist}
\end{figure*}

\noindent \textbf{Results:}
We present the results for discovering unknown classes on MNIST in \figref{fig:res_mnist}. We observe that the conditioning and targeting strategy as in \model\ using \textsc{scg+mi} and \textsc{scg+cmi} acquisition functions outperforms the uncertainty and diversity based methods by $\approx 5 - 15\%$ in terms of the average accuracy on the unknown classes (see \figref{fig:res_mnist} (a)). They obtain this gain in accuracy quickly, in the early rounds of AL and maintain it till the end. This is due to the fact that \model\ based strategies are able to select more data points from the unknown classes (see \figref{fig:res_mnist}(b,c,d)). Particularly, \textsc{Flcg+cmi} finds all the data points from the unknown classes in early rounds of AL. However, as we discussed in Sec \secref{sec:our_method}, \textsc{Flcmi} is computationally much more expensive than \textsc{Flmi}, and as we can see in \figref{fig:res_mnist}(a), \textsc{Flmi} eventually obtains the same accuracy in the later rounds of AL.

\begin{figure*}
\centering
\includegraphics[width = 14cm, height=1cm]{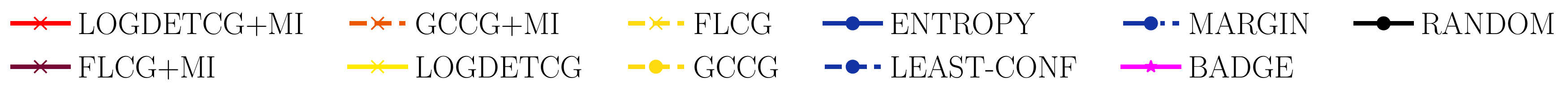}
\centering
\begin{subfigure}[t]{0.33\textwidth}
\includegraphics[width = \textwidth]{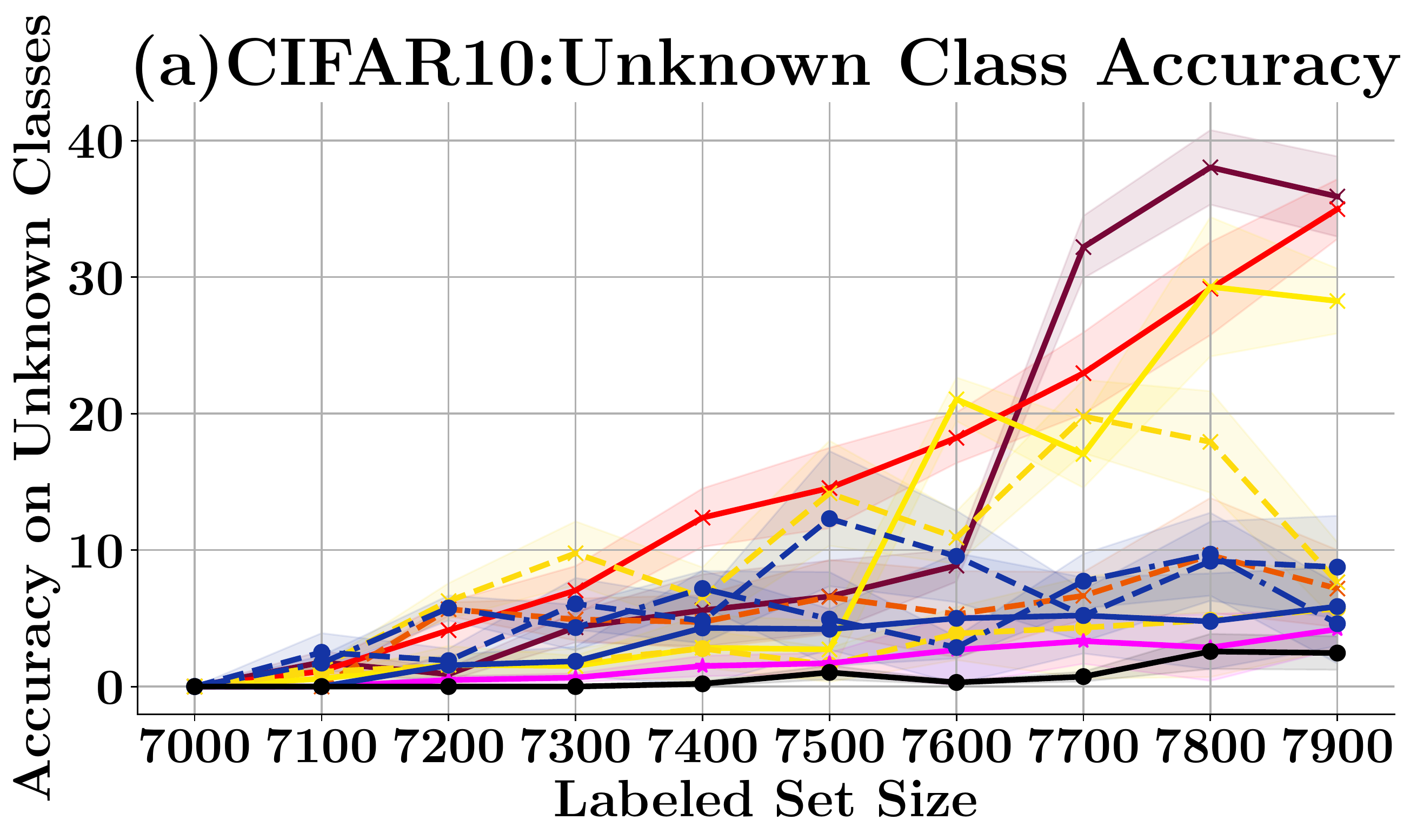}
\end{subfigure} 
\begin{subfigure}[t]{0.33\textwidth}
\includegraphics[width = \textwidth]{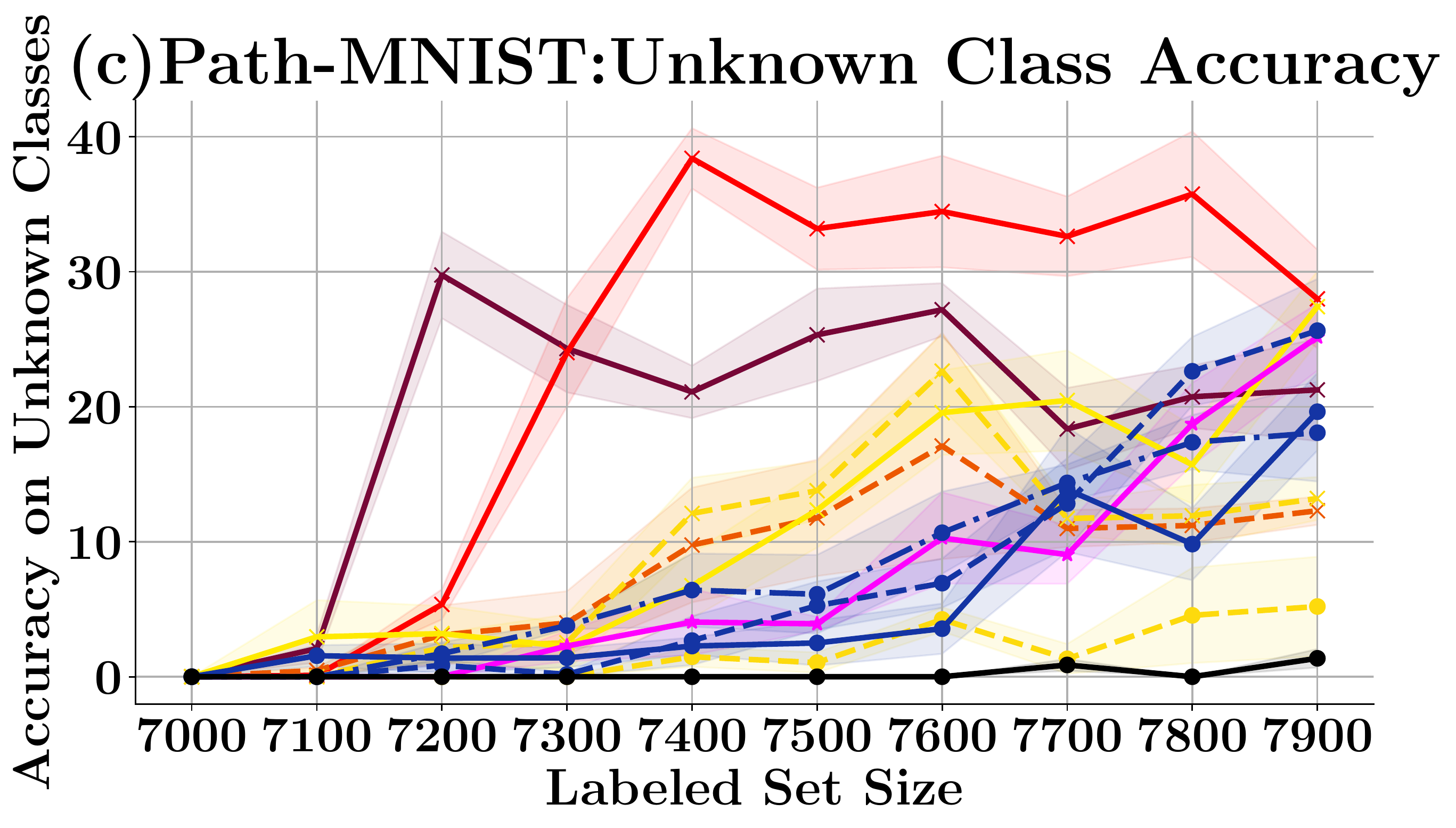}
\end{subfigure}
\begin{subfigure}[b]{0.325\textwidth}
\includegraphics[width = \textwidth]{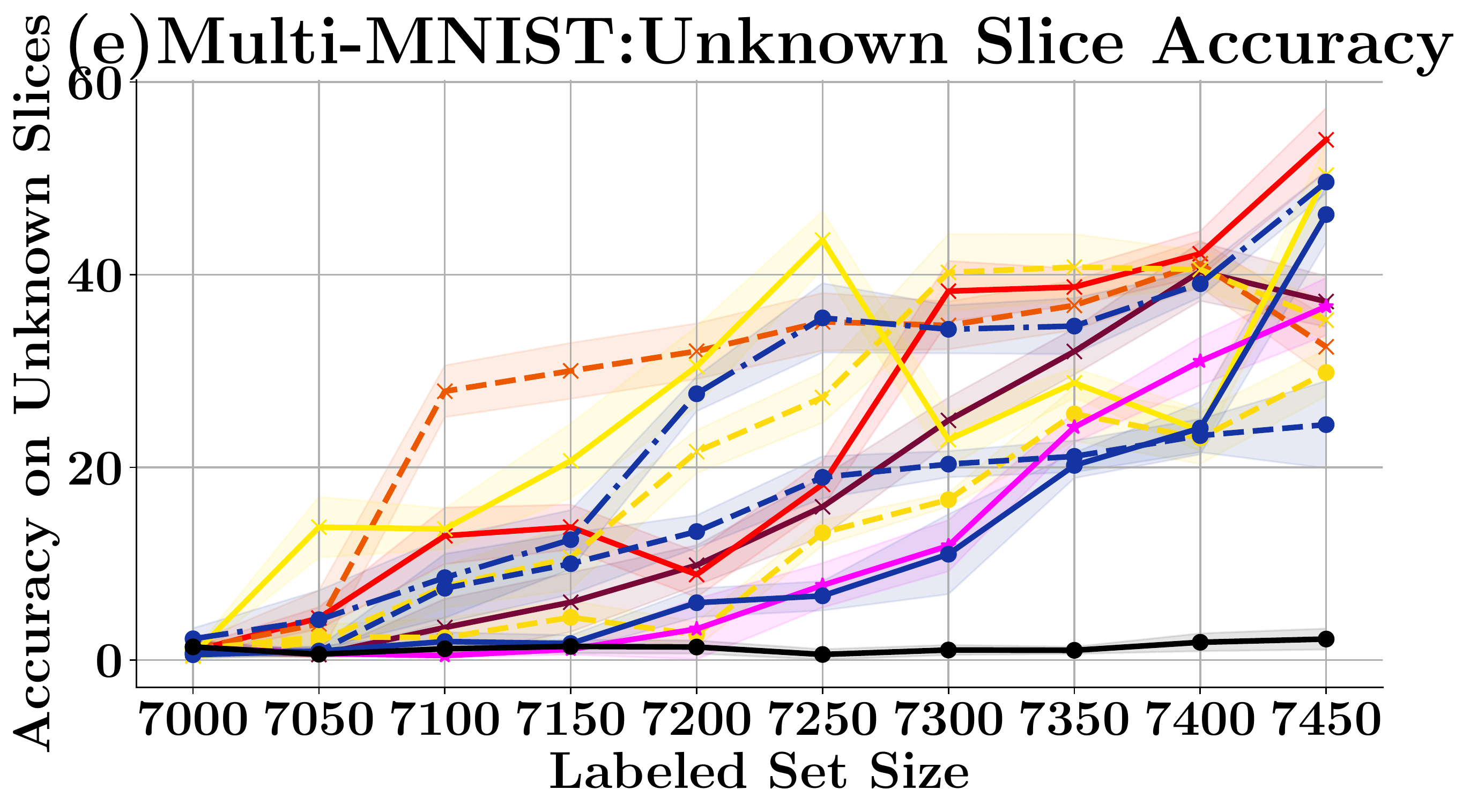}
\end{subfigure}

\begin{subfigure}[t]{0.33\textwidth}
\includegraphics[width = \textwidth]{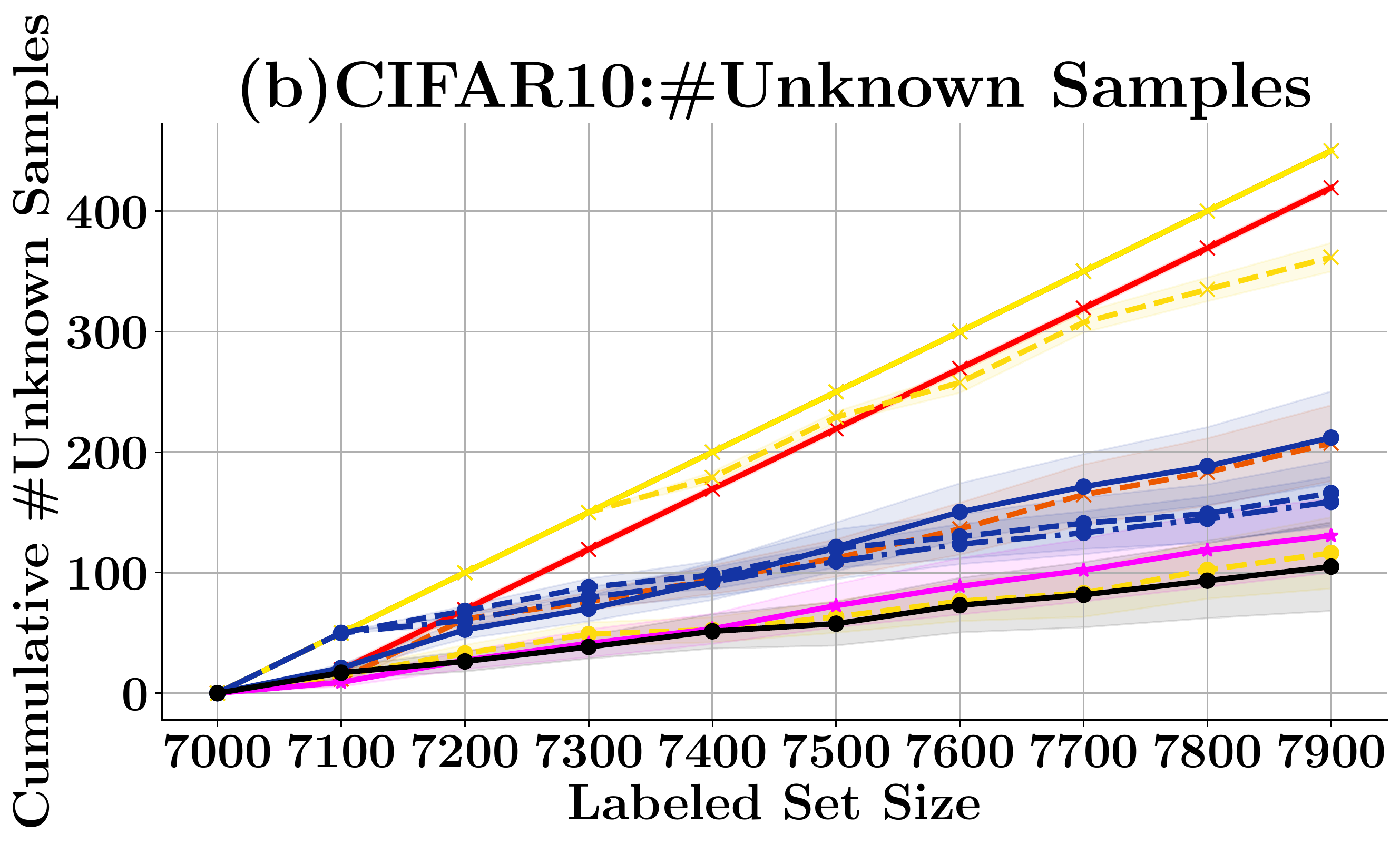}
\end{subfigure} 
\begin{subfigure}[t]{0.33\textwidth}
\includegraphics[width = \textwidth]{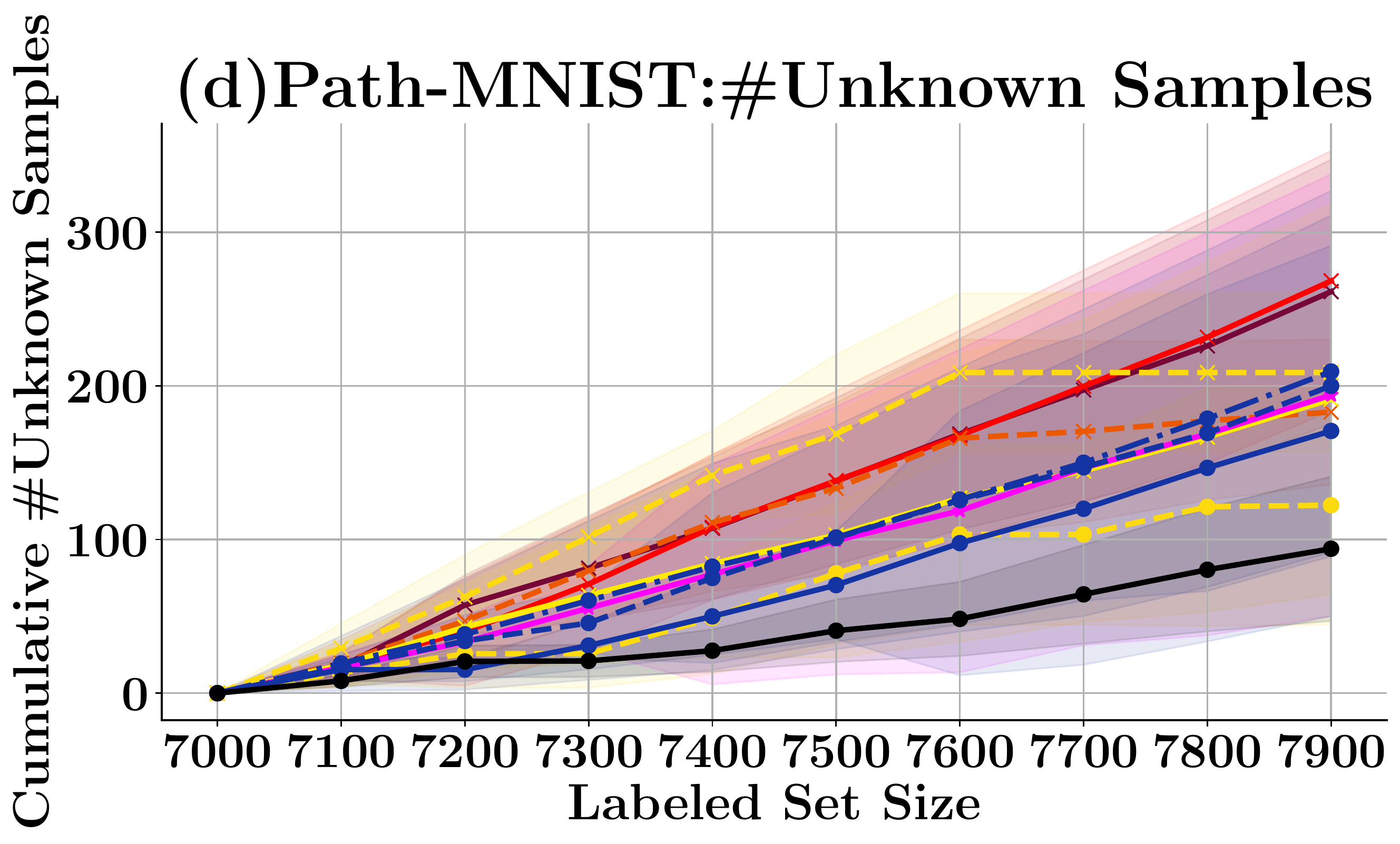}
\end{subfigure}
\begin{subfigure}[t]{0.32\textwidth}
\includegraphics[width = \textwidth]{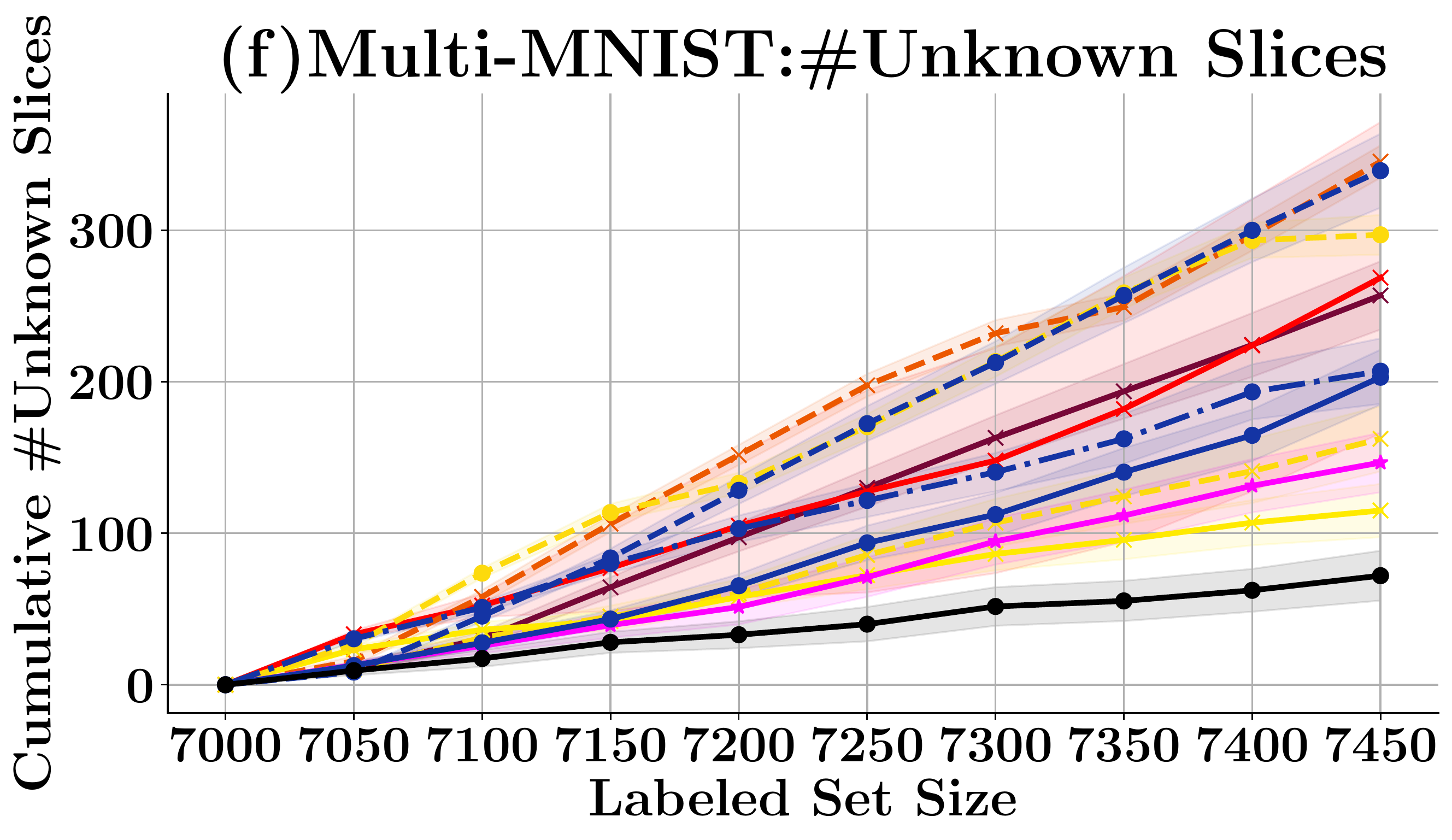}
\end{subfigure}
\caption{Active Learning for discovering unknown classes on CIFAR-10 \cite{krizhevsky2009learning} (\textbf{left} col.) and Path-MNIST \cite{medmnistv2, kather2019predicting}(\textbf{center col.}), and for discovering unknown slices on Multi-MNIST \cite{jiang2020mnist}(\textbf{right} col). \textsc{Flcg+mi} and \textsc{Logdetcg+mi} discovers the best subset of unknown class data points in early rounds of AL, thereby quickly gaining high accuracy for unknown classes. \textsc{Gccg+mi} discovers the best subset of unknown slices and obtains high accuracy for unknown slices.
\vspace{-3ex}}
\label{fig:res_cifar_path_multi}
\end{figure*}

In \figref{fig:res_cifar_path_multi}, we compare the computationally cheaper strategy of \model\ (\textsc{scg+mi}) for discovering unknown classes on two additional datasets - CIFAR10 (\figref{fig:res_cifar_path_multi}(a,b)) and Path-MNIST (\figref{fig:res_cifar_path_multi}(c,d)). We observe that the best performing methods are \textsc{scg+mi} variants which model representation (\textsc{Fl}) and diversity (\textsc{Logdet}) in addition to relevance. Particularly, \textsc{Flcg+mi} and \textsc{Logdetcg+mi} outperform other methods by $\approx 5 - 15\%$ in terms of the average accuracy on the unknown classes. Importantly, they acquire the best subset of unknown class data points in the early AL rounds (see \figref{fig:res_cifar_path_multi}(b,d)), thereby quickly reaching high accuracy values (see \figref{fig:res_cifar_path_multi}(a,c)).

\subsubsection{Unknown Slices} \label{sec:u_slices}

\noindent \textbf{Datasets and Experimental setup:}
For discovering rare slices, we apply our framework to Multi-MNIST, a dataset of images of digits from multiple languages. We consider two languages, English and Kannada, and try to train a single digit classification model for both the languages. In order to simulate unknown slices, we create a labeled set $\Lcal$ which is missing data points for a few digits \emph{only} for the Kannada language - 1,5, and 6 in our experiments. Note that $\Lcal$ contains data points for all digits for the English language. We refer to these missing digits from the Kannada language as the unknown slice of data. Since these Kannada digits are unknown in $\Lcal$, we create an imbalance for the Kannada language digits in the unlabeled set $\Ucal$, as done in \secref{sec:u_classes}. For the labeled set, we use $|\Lcal^{en}|=10K$, $|\Lcal^{ka}|=7K$, and for the unlabeled set , we use $|\Ucal^{en}|=10K$, $|\Ucal^{ka}|=7.6K$, an imbalance factor $\rho=5$, and a batch size $B=50$. The superscripts \textit{en} and \textit{ka} denote the English and Kannada data slices, respectively.

\noindent \textbf{Results:}
We present the results for discovering rare slices in \figref{fig:res_cifar_path_multi}. We observe that functions that model relevance to the query set $\Qcal$ outperform other functions in discovering rare slices. Particularly, \textsc{Gccg+mi} followed by \textsc{Logdet+mi} outperforms other methods by $\approx 5 - 15\%$ in terms of average accuracy on the rare slices (see \figref{fig:res_cifar_path_multi}(e)). This is due to the fact that \textsc{Gccg+mi} selects the most number of data points from the unknown slices (see \figref{fig:res_cifar_path_multi}(f)).

\begin{figure*}
\centering
\includegraphics[width = 14cm]{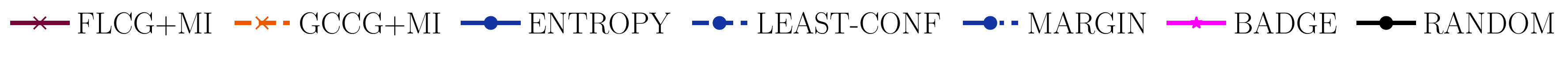}
\begin{subfigure}[t]{0.33\textwidth}
\includegraphics[width = \textwidth]{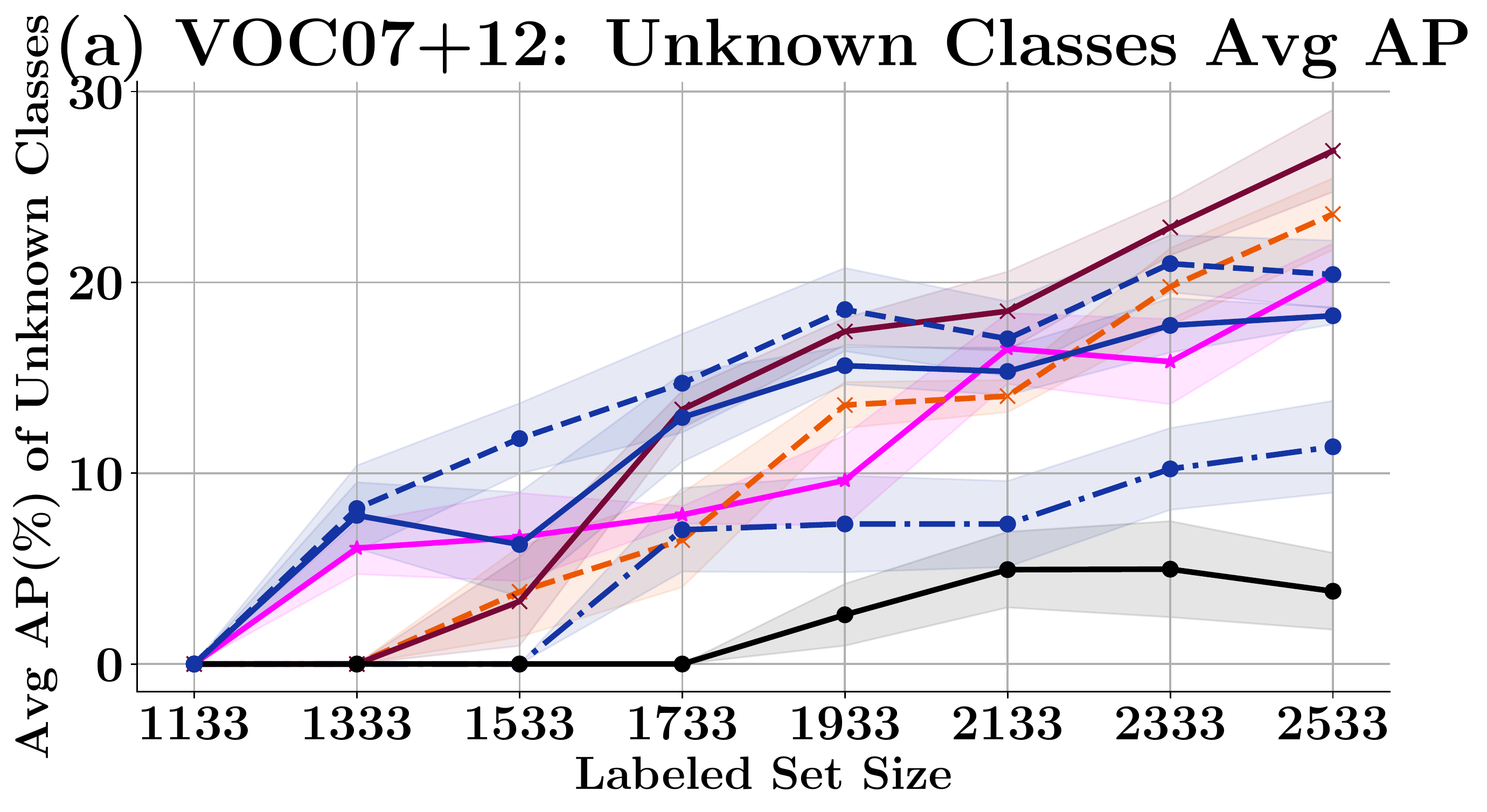}
\end{subfigure}
\begin{subfigure}[t]{0.33\textwidth}
\includegraphics[width = \textwidth]{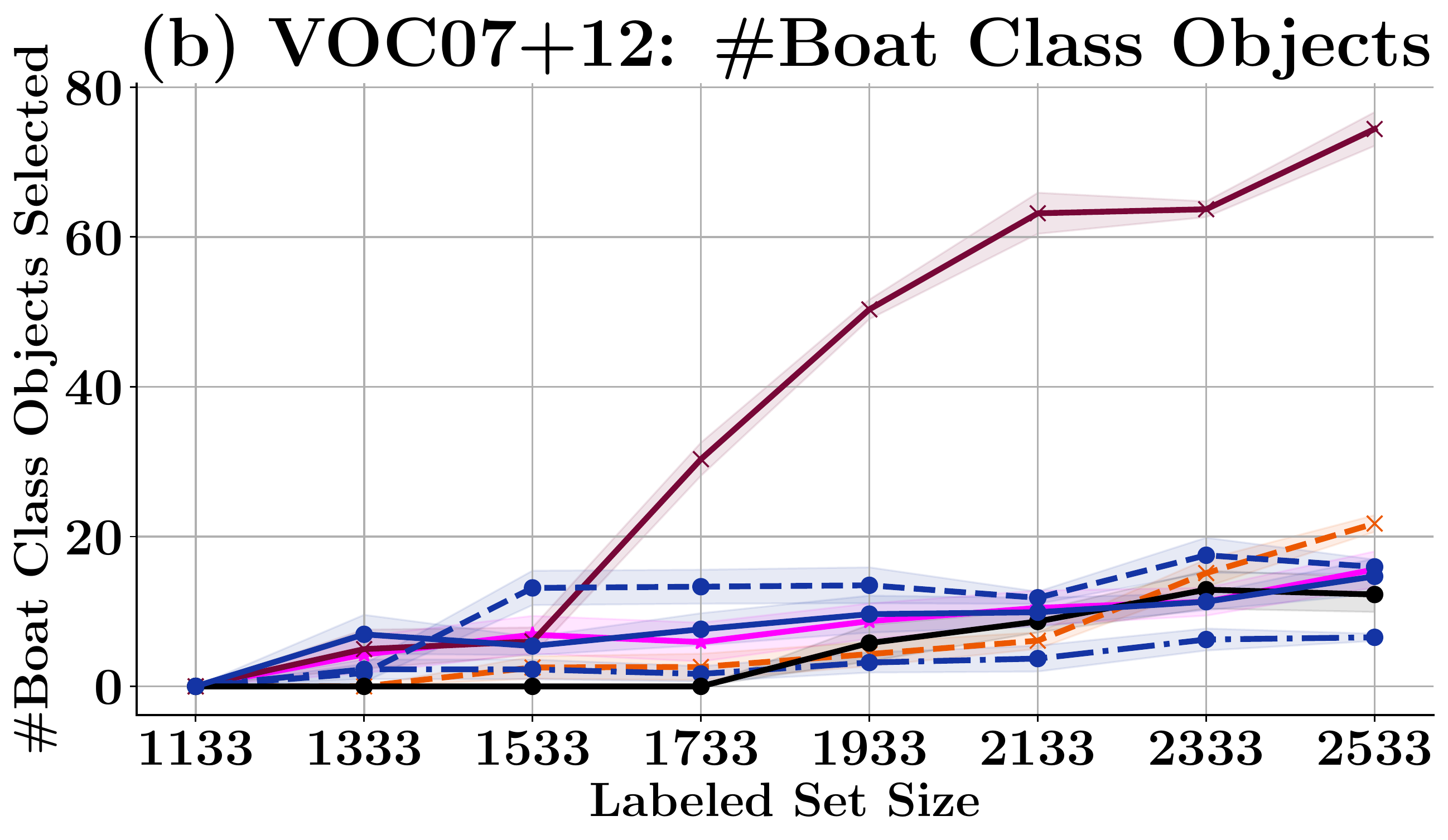}
\end{subfigure}
\begin{subfigure}[t]{0.325\textwidth}
\includegraphics[width = \textwidth]{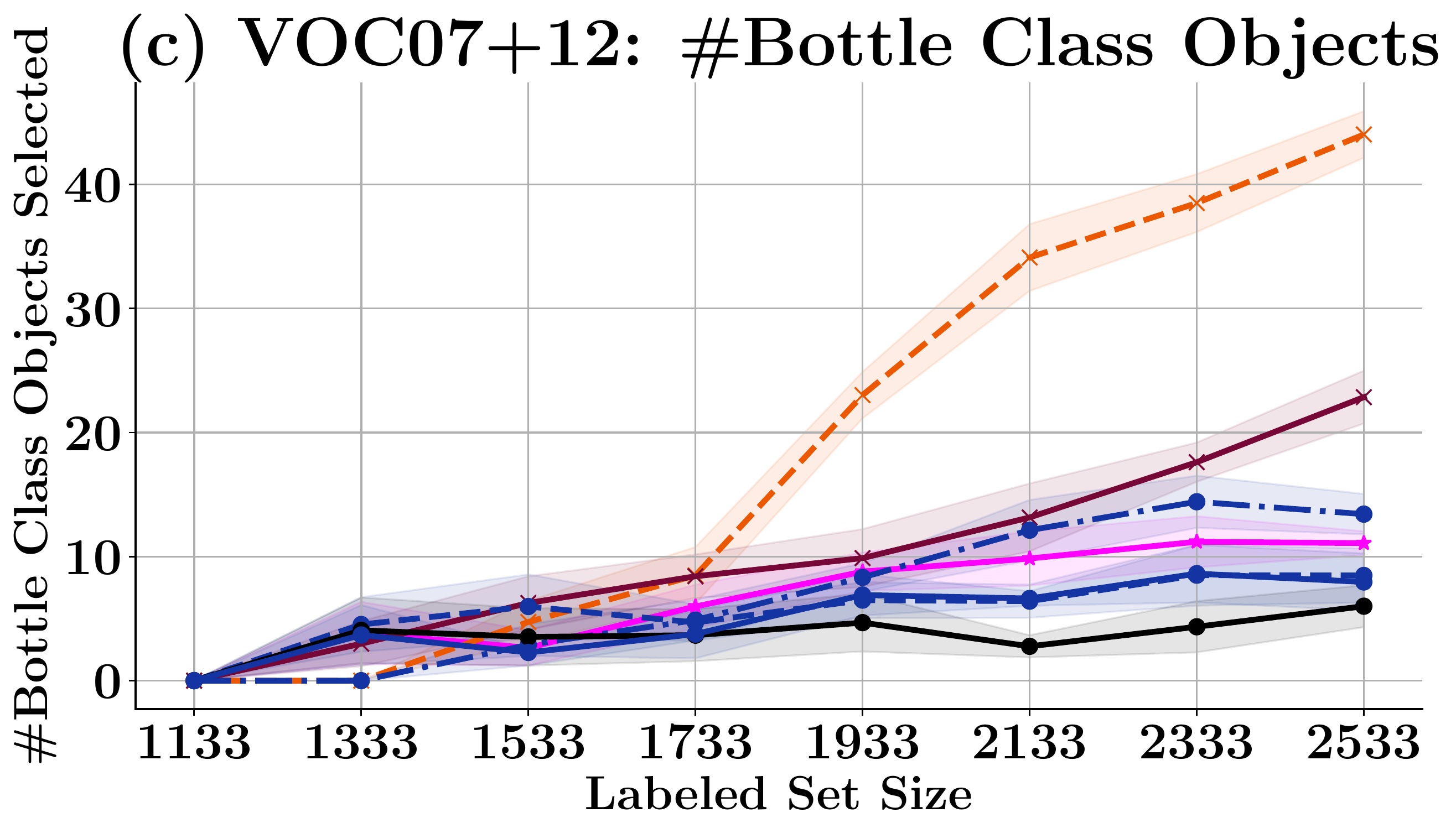}
\end{subfigure}
\caption{Active Data Discovery for unknown classes on PASCAL-VOC. Plot (a) shows the average AP of unknown classes on the test set while plot (b, c) shows the number of data points discovered from unknown classes of boat and bottle. We observe that the \textsc{Scg+mi} (\textsc{Flcg+mi} and \textsc{Gccg+mi}) outperform other methods in terms of the average AP on the unknown classes by $\approx 5\% - 7\%$. \textsc{Scg+mi} also selects maximum data points from both the unknown class objects.
\vspace{-3ex}
}
\label{fig:res_voc}
\end{figure*}

\subsection{Object Detection} \label{sec:exp_od}
In this section, we present the results of active data discovery in the context of object detection tasks. We evaluate the performance of \model\ with existing AL acquisition functions for discovering unknown classes by comparing (1) the number of unknown objects selected by each acquisition function, and (2) average AP of unknown classes evaluated on the test set. For all experiments, we train a Faster RCNN model~\cite{ren2015faster} with a ResNet50 backbone~\cite{he2016deep}. In each AL round, we reinitialize the model parameters and train for 150 epochs using SGD with momentum. We set the initial learning rate and step size to 0.001 and 3, momentum and weight decay to 0.9 and 0.0005.

\noindent \textbf{Datasets and Experimental setup:} 
We ran our data discovery experiments for object detection on the well known public dataset PASCAL VOC07+12 (VOC07+12) \cite{everingham2010pascal} 
Each image in VOC07+12 can contain multiple objects of different classes. 
We create the initial labeled set, $\mathcal{L}$ to simulate the unknown classes scenario by creating a class imbalance at an object level. Let $\mathcal{C}_i^\mathcal{L}$ be the number of objects from an unknown (target) class $i$ and $\mathcal{D}_j^\mathcal{L}$ be the number of objects from a known class $j$. The initial labeled set $\mathcal{L}$ is created such that $\mathcal{C}_i^\mathcal{L} = 0\,$ i.e. no unknown class images at the beginning and $\mathcal{D}_j^\mathcal{L} = 100\,$ i.e. at least 100 objects of each known class to start with. This gives us an initial labeled set of size $|\mathcal{L}|=1133$ images. Note that the imbalance is not exact because objects of some known classes like person are predominantly present in most images, thereby increasing the size of $|\mathcal{D}_j^\mathcal{L}|$ for those classes. To check the robustness of our model on extreme unknown scenarios, we constrain the unlabeled set $\mathcal{U}$ to contain maximum of $150$ objects of the unknown classes each. We choose two classes `boat' and `bottle' from VOC07+12 to be the \textit{unknowns} since they are (1) one of the most uncommon classes in VOC, thereby making them hard to discover, and (2) they are or appear as comparatively smaller objects in the images than other classes which are either big or occupy most of the foreground. \looseness-1

\noindent \textbf{Results:} In \figref{fig:res_voc}, we compare the performance of different AL strategies for discovering unknown objects in VOC07+12. We observe that our proposed \scg\ + \smi\ methods, particularly (\textsc{FLcg+mi} and \textsc{Gccg+mi}) surpass other baselines in terms of the average AP of unknown classes and the total unknown samples discovered from the unlabeled dataset. We do not compare with the \scmi\ and log determinant methods, since we find them intractable for object detection tasks due to the large number of proposals for each image and the complexity of log determinant.

\vspace{-2ex}
\section{Conclusion} \label{sec:concl}

In this paper, we propose \model\, an active learning based framework for data discovery. We show the effectiveness of \model\ for image classification and object detection tasks in discovery of unknown instances across classes and slices for a wide range of diverse datasets. Our experiments show that using a combination of \scg\ and \smi\ is the most effective and scalable, and obtains a $\approx 5\% - 15\%$ gain compared to existing baselines. The main limitation of this work is that the known instances need to be well represented in the feature space so that they are not mixed with the unknown instances. A potential negative societal impact of \model\ is that it can be used to discover and incorporate new biases in the dataset.

\bibliographystyle{splncs04}
\bibliography{main}

\newpage

\appendix

\setcounter{page}{1}

\newpage

\begin{center}
\part*{Supplementary Material for Active Data Discovery: Mining Unknown Data using Submodular Information Measures} 
\end{center}

\tableofcontents

\section{Summary of Notations}\label{app:notation-summary}

 \begin{table*}[!h]
 \centering
 \begin{tabular}{|l|l|p{0.5\textwidth}|} 
 \toprule
 \hline 
 \multicolumn{1}{|l|}{Topic} & Notation & Explanation \\ \hline
 \toprule \hline
 &  $\Ucal$ & Unlabeled set of $|\Ucal|$ instances\\ 
 \multicolumn{1}{|p{0.20\textwidth}|}{\model\ (Sec. 3)} 
 & $\Acal$ & A subset of $\Ucal$\\ 
 & $S_{ij}$ & Similarity between any two data points $i$ and $j$\\
 & $f$ & A submodular function\\
 & $\Lcal$ & Labeled set of data points\\
 & $\Qcal$ & Query set\\
 & $\Pcal$ & Private set\\
 & $\Acal^\Qcal$ & Subset of the selected subset containing data points from the \emph{unknown} classes\\
 & $\Acal^\Pcal$ & Subset of the selected subset containing data points from the \emph{known} classes\\
 & $\Mcal$ & Deep model\\
 & $B$ & Active learning selection budget\\
 & $\Kcal$ & Set of unique known class IDs\\
 & $\Acal^C$ & Set of class IDs from the newly selected subset $\Acal$\\
 & $\Scal^\Pcal$ & Similarity kernel between the unlabeled set $\Ucal$ and the set with known class data points $\Pcal$\\
 & $\Scal^\Qcal$ & Similarity kernel between the unlabeled set $\Ucal$ and the set with unknown class data points $\Qcal$\\
 \hline
 \bottomrule
 \end{tabular}
 \caption{Summary of notations used throughout this paper}
 \label{tab:main-notations}
 \end{table*}
 



\section{More Details on Experimental Setup, Datasets, and Baselines}
 
\subsection{Model Architecture and Hyperparameters} \label{app:hyper_params}

We used a standard Resnet18 \cite{resnet} architecture as the backbone for all the experiments performed across the datasets. The hyperparamters used for the various experiments are summarised in Table \ref{tab:hyper_params_table}. All the experiments start with zero training samples of the unknown classes and subsequently newly found samples known and unknown are added from the Query set into the private and query set as well as the training set.

\begin{table*}[!h]
\small
 \centering
 \begin{tabular}{|l|l|l|l|l|l|l|l|} 
 \toprule
 \hline 
 \multicolumn{1}{|l|}{Dataset} & Model & \multicolumn{6}{|l|}{Hyperparameters} \\
 \cline{3-8}
& & Learning Rate & Batch Size & Epochs &
\multicolumn{3}{|l|}{Per Class Splits} \\
\cline{6-8}
& & & & & Train & Test & Lake \\
\hline
\multirow{2}{*}{MNIST} & \multirow{2}{*}{Resnet18} & \multirow{2}{*}{0.0005} & \multirow{2}{*}{32} & \multirow{2}{*}{200} & Known: 1000 & \multirow{2}{*}{100} & Known: 1000 \\
 & & & & & Unknown: 0 & & Unknown: 50 \\
 \hline
 \multirow{2}{*}{CIFAR10} & \multirow{2}{*}{Resnet18} & \multirow{2}{*}{0.001} & \multirow{2}{*}{32} & \multirow{2}{*}{200} & Known: 1000 & \multirow{2}{*}{100} & Known: 1000 \\
 & & & & & Unknown: 0 & & Unknown: 300 \\
 \hline
 \multirow{2}{*}{Path-MNIST} & \multirow{2}{*}{Resnet18} & \multirow{2}{*}{0.0005} & \multirow{2}{*}{32} & \multirow{2}{*}{200} & Known: 1000 & \multirow{2}{*}{100} & Known: 1000 \\
 & & & & & Unknown: 0 & & Unknown: 300 \\
 \hline
 \multirow{2}{*}{Multi-MNIST} & \multirow{2}{*}{Resnet18} & \multirow{2}{*}{0.0005} & \multirow{2}{*}{32} & \multirow{2}{*}{200} & Known: 500 & \multirow{2}{*}{100} & Known: 1000 \\
 & & & & & Unknown: 0 & & Unknown: 200 \\
 \hline
 \bottomrule
 \end{tabular}
 \caption{Details of hyperparameters used for each experiment}
 \label{tab:hyper_params_table}
 \end{table*}
 
\subsection{Strategy Configurations} \label{tab:strategy_config}
Extensive experiments were performed to optimise the configurations for each strategy on the various datasets. The configurations with the best reproducible results are listed for each of the datasets in Tables \ref{tab:strategy_mnist}, \ref{tab:strategy_cifar}, \ref{tab:strategy_path}, \ref{tab:strategy_multi}. 

 \begin{table}[!h]
 \centering
 \begin{tabular}{|l|l|l|l|l|l|l|l|} 
 \toprule
 \hline 
 \multirow{3}{*}{Dataset} & \multirow{3}{*}{Strategy} &  \multicolumn{5}{|l|}{Strategy Configurations} \\ 
\cline{3-7}
 
 & & Unknown & \multirow{2}{*}{Embedding Type} & \multirow{2}{*}{$\nu$} & \multirow{2}{*}{$\eta$} & \multirow{2}{*}{Budget} \\
 & & Classes &  &  & &  \\
 \toprule \hline
 \multirow{17}{*}{MNIST} & LOGDETCG + MI & 3 & Features & 1.0 & 1.0 & 50 \\
 \cline{2-7}
 & FLCG + MI & 3 & \multirow{2}{*}{Features} & FLCG: 1.5 & 1.0 & 50 \\
 &  &  &  & MI: 1.0 & 1.0 & 50 \\
 \cline{2-7}
 & GCCG + MI & 3 & \multirow{2}{*}{Features} & GCCG: 1.5 & 1.0 & 50 \\
 &  &  &  & MI: 1.0 & 1.0 & 50 \\
 \cline{2-7}
 & FLCG + CMI & 3 & \multirow{2}{*}{Features} & FLCG: 1.5 & 1.0 & 50 \\
 &  &  &  & MI: 1.0 & 1.0 & 50 \\
 \cline{2-7}
 & GCCG + CMI & 3 & \multirow{2}{*}{Features} & GCCG: 1.5 & 1.0 & 50 \\
 &  &  &  & MI: 1.0 & 1.0 & 50 \\
 \cline{2-7}
 & LOGDETCG & 3 & Features & 1.0 & 1.0 & 50 \\
 \cline{2-7}
 & FLCG & 3 & Features & 1.7 & 1.0 & 50 \\
 \cline{2-7}
 & GCCG & 3 & Features & 1.7 & 1.0 & 50 \\
 \cline{2-7}
 & Entropy & 3 & Class scores & - & - & 50 \\
 \cline{2-7}
 & Least Confidence & 3 & Class scores & - & - & 50 \\
 \cline{2-7}
 & Margin & 3 & Class scores & - & - & 50 \\
 \cline{2-7}
 & Badge & 3 & Gradients & - & - & 50 \\
 \cline{2-7}
 & Random & 3 & - & - & - & 50 \\
 \hline
 \bottomrule
 \end{tabular}
 \caption{Summary of strategy configurations used for MNIST dataset.}
 \label{tab:strategy_mnist}
 \end{table}

 \begin{table}[!h]
 \centering
 \begin{tabular}{|l|l|l|l|l|l|l|l|} 
 \toprule
 \hline 
 \multirow{3}{*}{Dataset} & \multirow{3}{*}{Strategy} &  \multicolumn{5}{|l|}{Strategy Configurations} \\ 
\cline{3-7}
 
 & & Unknown & \multirow{2}{*}{Embedding Type} & \multirow{2}{*}{$\nu$} & \multirow{2}{*}{$\eta$} & \multirow{2}{*}{Budget} \\
 & & Classes &  &  & &  \\
 \toprule \hline
\multirow{17}{*}{CIFAR10} & LOGDETCG + MI & 3 & Features & 1.0 & 1.0 & 100 \\
 \cline{2-7}
 & FLCG + MI & 3 & \multirow{2}{*}{Features} & FLCG: 1.5 & 1.0 & 100 \\
 &  &  &  & MI: 1.0 & 1.0 & 100 \\
 \cline{2-7}
 & GCCG + MI & 3 & \multirow{2}{*}{Features} & GCCG: 1.5 & 1.0 & 100 \\
 &  &  &  & MI: 1.0 & 1.0 & 100 \\
 \cline{2-7}
 & FLCG + CMI & 3 & \multirow{2}{*}{Features} & FLCG: 1.5 & 1.0 & 100 \\
 &  &  &  & MI: 1.0 & 1.0 & 100 \\
 \cline{2-7}
 & GCCG + CMI & 3 & \multirow{2}{*}{Features} & GCCG: 1.5 & 1.0 & 100 \\
 &  &  &  & MI: 1.0 & 1.0 & 100 \\
 \cline{2-7}
 & LOGDETCG & 3 & Features & 1.0 & 1.0 & 100 \\
 \cline{2-7}
 & FLCG & 3 & Features & 1.7 & 1.0 & 100 \\
 \cline{2-7}
 & GCCG & 3 & Features & 1.7 & 1.0 & 100 \\
 \cline{2-7}
 & Entropy & 3 & Class scores & - & - & 100 \\
 \cline{2-7}
 & Least Confidence & 3 & Class scores & - & - & 100 \\
 \cline{2-7}
 & Margin & 3 & Class scores & - & - & 100 \\
 \cline{2-7}
 & Badge & 3 & Gradients & - & - & 100 \\
 \cline{2-7}
 & Random & 3 & - & - & - & 100 \\
 \hline
 \bottomrule
 \end{tabular}
 \caption{Summary of strategy configurations used for CIFAR10 dataset.}
 \label{tab:strategy_cifar}
 \end{table}

 \begin{table}[!h]
 \centering
 \begin{tabular}{|l|l|l|l|l|l|l|l|} 
 \toprule
 \hline 
 \multirow{3}{*}{Dataset} & \multirow{3}{*}{Strategy} &  \multicolumn{5}{|l|}{Strategy Configurations} \\ 
\cline{3-7}
 
 & & Unknown & \multirow{2}{*}{Embedding Type} & \multirow{2}{*}{$\nu$} & \multirow{2}{*}{$\eta$} & \multirow{2}{*}{Budget} \\
 & & Classes &  &  & &  \\
 \toprule \hline
\multirow{17}{*}{Path-MNIST} & LOGDETCG + MI & 3 & Features & 1.0 & 1.0 & 100 \\
 \cline{2-7}
 & FLCG + MI & 3 & \multirow{2}{*}{Features} & FLCG: 1.5 & 1.0 & 100 \\
 &  &  &  & MI: 1.0 & 1.0 & 100 \\
 \cline{2-7}
 & GCCG + MI & 3 & \multirow{2}{*}{Features} & GCCG: 1.5 & 1.0 & 100 \\
 &  &  &  & MI: 1.0 & 1.0 & 100 \\
 \cline{2-7}
 & FLCG + CMI & 3 & \multirow{2}{*}{Features} & FLCG: 1.5 & 1.0 & 100 \\
 &  &  &  & MI: 1.0 & 1.0 & 100 \\
 \cline{2-7}
 & GCCG + CMI & 3 & \multirow{2}{*}{Features} & GCCG: 1.5 & 1.0 & 100 \\
 &  &  &  & MI: 1.0 & 1.0 & 100 \\
 \cline{2-7}
 & LOGDETCG & 3 & Features & 1.0 & 1.0 & 100 \\
 \cline{2-7}
 & FLCG & 3 & Features & 1.7 & 1.0 & 100 \\
 \cline{2-7}
 & GCCG & 3 & Features & 1.7 & 1.0 & 100 \\
 \cline{2-7}
 & Entropy & 3 & Gradients & - & - & 100 \\
 \cline{2-7}
 & Least Confidence & 3 & Gradients & - & - & 100 \\
 \cline{2-7}
 & Margin & 3 & Gradients & - & - & 100 \\
 \cline{2-7}
 & Badge & 3 & Gradients & - & - & 100 \\
 \cline{2-7}
 & Random & 3 & - & - & - & 100 \\
 \hline
 \bottomrule
 \end{tabular}
 \caption{Summary of strategy configurations used for Path-MNIST dataset.}
 \label{tab:strategy_path}
 \end{table}

 \begin{table}[!h]
 \centering
 \begin{tabular}{|l|l|l|l|l|l|l|l|} 
 \toprule
 \hline 
 \multirow{3}{*}{Dataset} & \multirow{3}{*}{Strategy} &  \multicolumn{5}{|l|}{Strategy Configurations} \\ 
\cline{3-7}
 
 & & Unknown & \multirow{2}{*}{Embedding Type} & \multirow{2}{*}{$\nu$} & \multirow{2}{*}{$\eta$} & \multirow{2}{*}{Budget} \\
 & & Classes &  &  & &  \\
 \toprule \hline
 \multirow{17}{*}{Multi-MNIST} & LOGDETCG + MI & 3 & Features & 1.0 & 1.0 & 50 \\
 \cline{2-7}
 & FLCG + MI & 3 & \multirow{2}{*}{Features} & FLCG: 1.5 & 1.0 & 50 \\
 &  &  &  & MI: 1.0 & 1.0 & 50 \\
 \cline{2-7}
 & GCCG + MI & 3 & \multirow{2}{*}{Features} & GCCG: 1.5 & 1.0 & 50 \\
 &  &  &  & MI: 1.0 & 1.0 & 50 \\
 \cline{2-7}
 & FLCG + CMI & 3 & \multirow{2}{*}{Features} & FLCG: 1.5 & 1.0 & 50 \\
 &  &  &  & MI: 1.0 & 1.0 & 50 \\
 \cline{2-7}
 & GCCG + CMI & 3 & \multirow{2}{*}{Features} & GCCG: 1.5 & 1.0 & 50 \\
 &  &  &  & MI: 1.0 & 1.0 & 50 \\
 \cline{2-7}
 & LOGDETCG & 3 & Features & 1.0 & 1.0 & 50 \\
 \cline{2-7}
 & FLCG & 3 & Features & 1.7 & 1.0 & 50 \\
 \cline{2-7}
 & GCCG & 3 & Features & 1.7 & 1.0 & 50 \\
 \cline{2-7}
 & Entropy & 3 & Gradients & - & - & 50 \\
 \cline{2-7}
 & Least Confidence & 3 & Gradients & - & - & 50 \\
 \cline{2-7}
 & Margin & 3 & Gradients & - & - & 50 \\
 \cline{2-7}
 & Badge & 3 & Gradients & - & - & 50 \\
 \cline{2-7}
 & Random & 3 & - & - & - & 50 \\
 \hline
 \bottomrule
 \end{tabular}
 \caption{Summary of strategy configurations used for Multi-MNIST dataset.}
 \label{tab:strategy_multi}
 \end{table}
 
\subsection{Licensing Details}
\paragraph{Datasets.} Our experiments with \textsc{ADD} utilize the following datasets.

\begin{itemize}
    \item \href{https://www.cs.toronto.edu/~kriz/cifar.html}{{\color{blue}CIFAR-10}}~\cite{krizhevsky2009learning}: MIT License
    \item \href{http://yann.lecun.com/exdb/mnist/}{{\color{blue}MNIST}}~\cite{lecun2010mnist}: Creative Commons Attribution-Share Alike 3.0
    \item \href{https://medmnist.com/}{{\color{blue}Path-MNIST}}~\cite{medmnistv2, kather2019predicting}: Creative Commons Attribution 4.0 International
    \item \href{https://github.com/jwwthu/MNIST-MIX}{{\color{blue}Multi-MNIST}}~\cite{jiang2020mnist}: Custom (Research, Non-Commercial)
\end{itemize}




\begin{figure*}[h]
\centering
\includegraphics[width = 14cm, height=1cm]{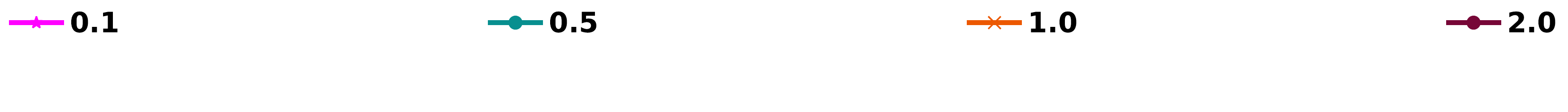}
\centering
\hspace*{-0.6cm}
\begin{subfigure}[t]{0.24\textwidth}
\includegraphics[width = \textwidth]{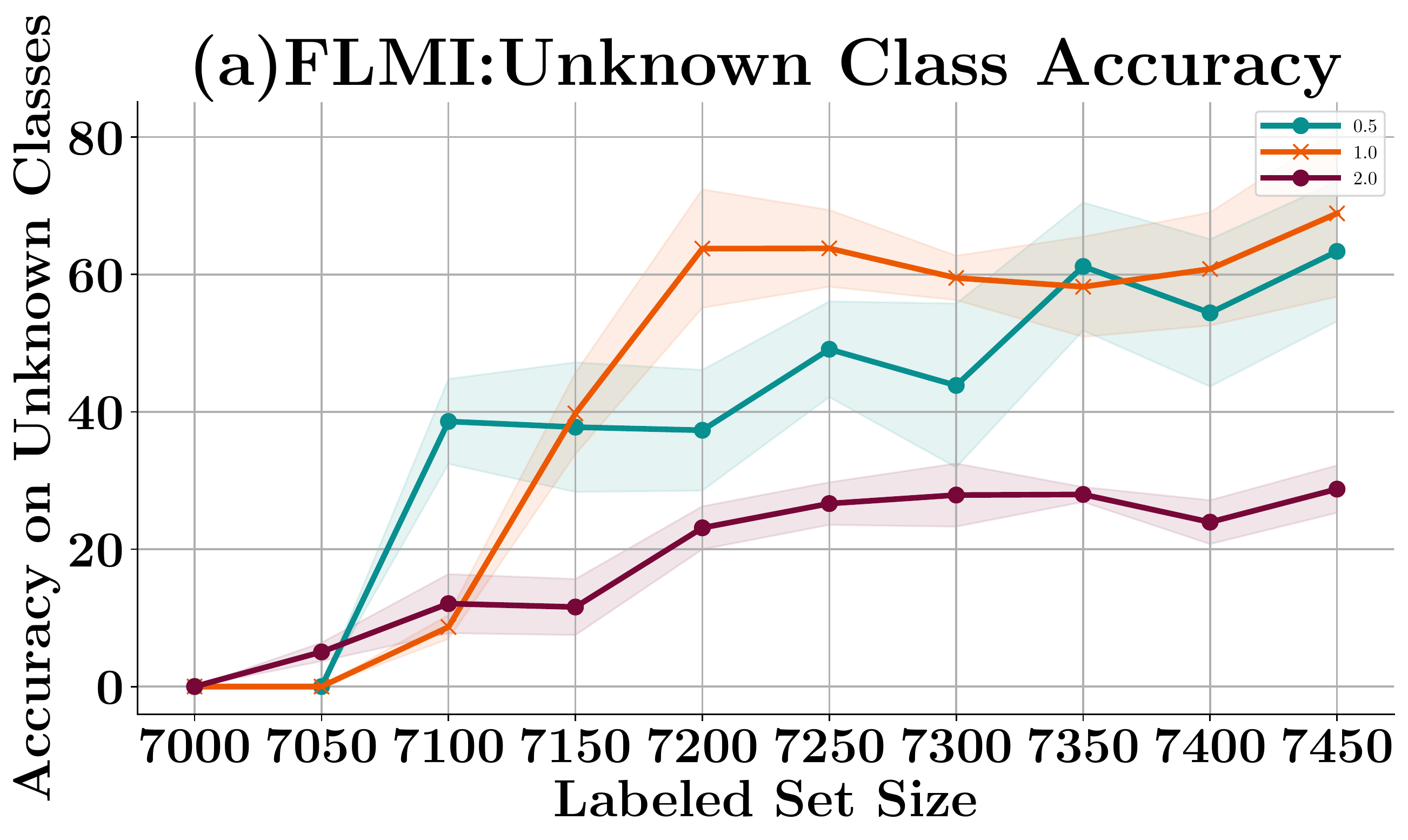}
\end{subfigure}
\begin{subfigure}[t]{0.25\textwidth}
\includegraphics[width = \textwidth]{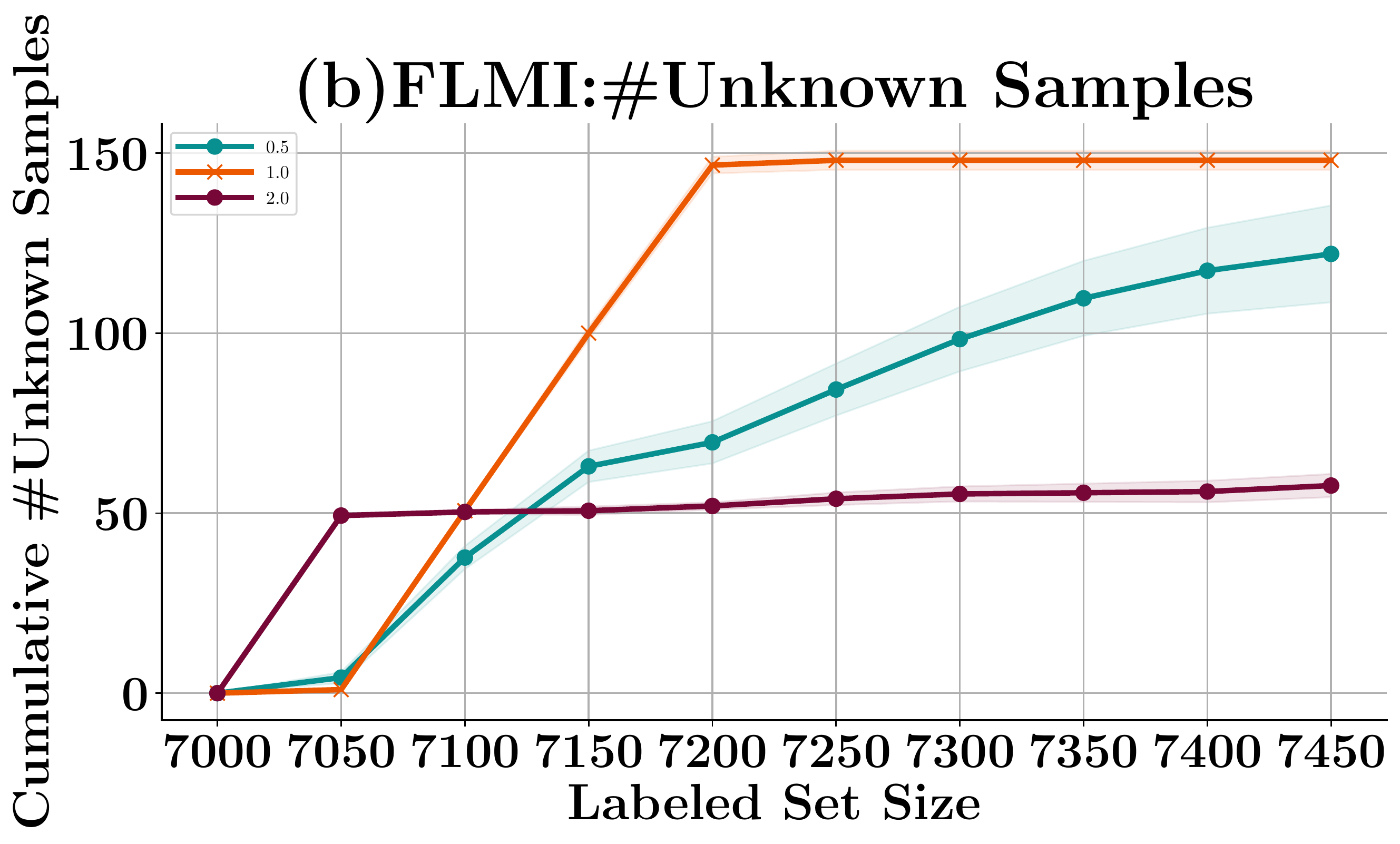}
\end{subfigure}
\begin{subfigure}[t]{0.25\textwidth}
\includegraphics[width = \textwidth]{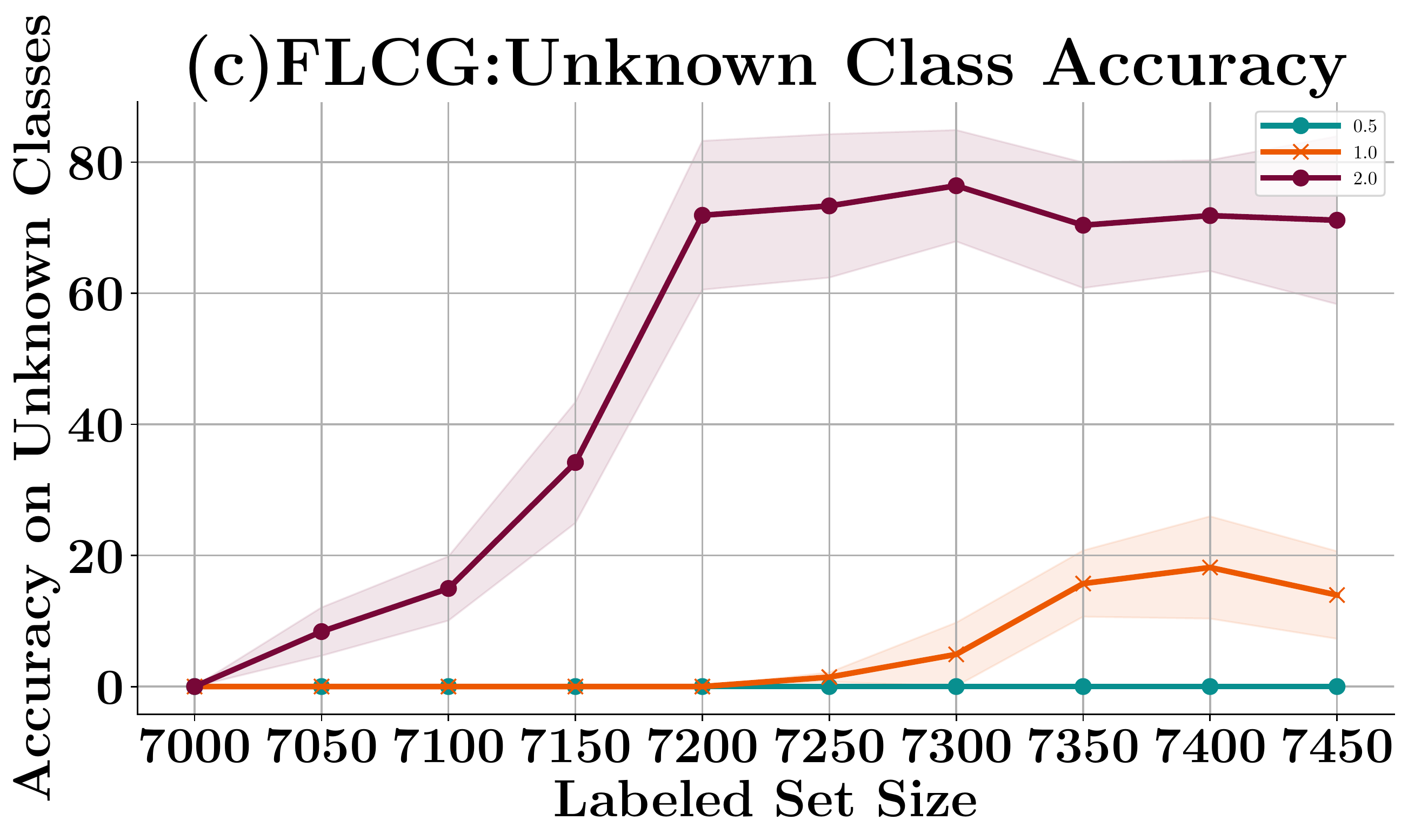}
\end{subfigure}
\begin{subfigure}[t]{0.25\textwidth}
\includegraphics[width = \textwidth]{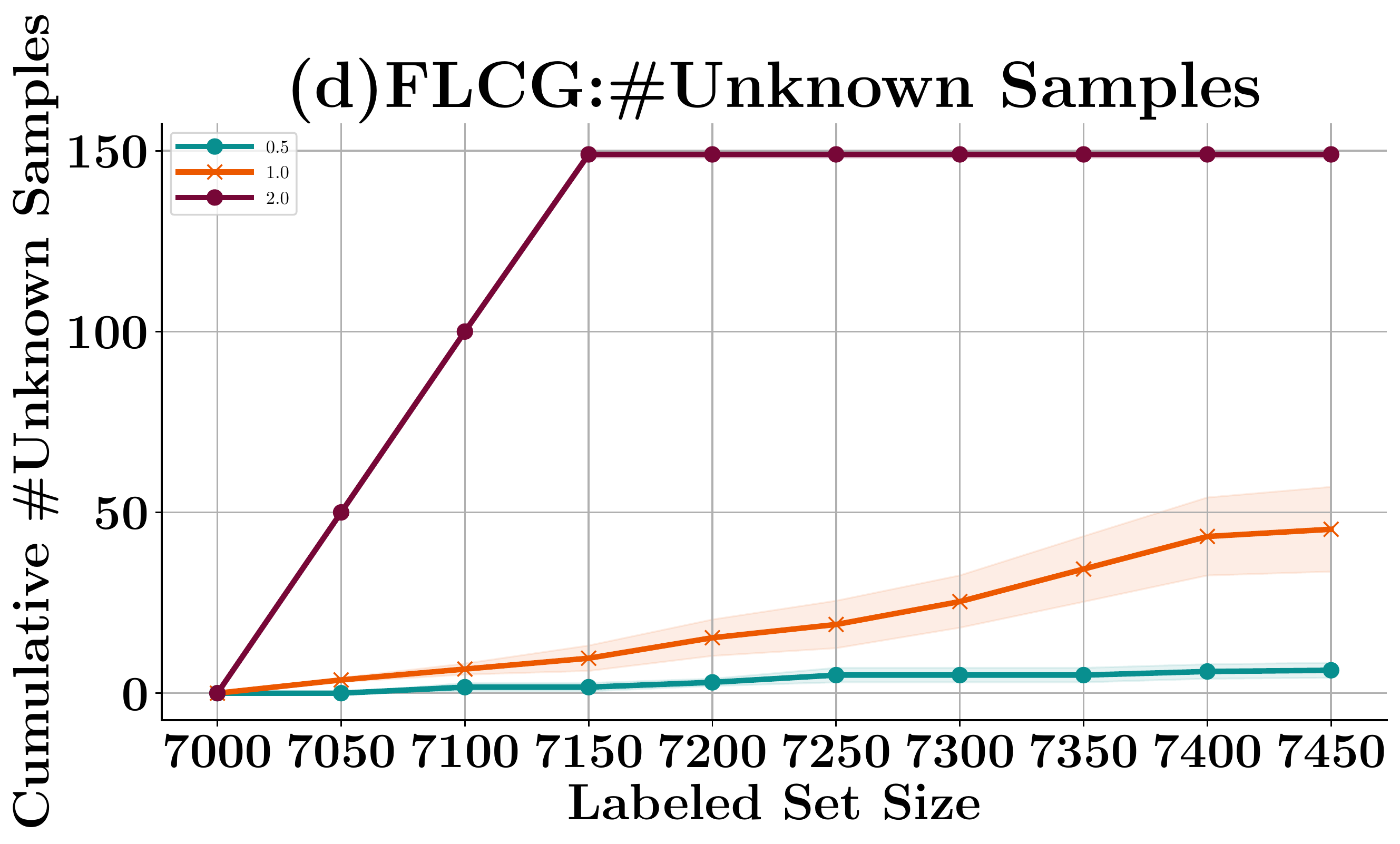}
\end{subfigure}
\caption{Ablation study demonstrating the affects of varying the conditioning and targeting variables on the discovery of unknown classes.
\vspace{-1ex}
}
\label{fig:res_ablation}
\end{figure*}

\section{Additional experiments} \label{app:add_exp}
\subsection{Ablation Study:} \label{sec:ablation}
We conduct an ablation study to analyze the effect of using various values of $\eta$ and $\nu$ for active data discovery. We present the results of this ablation study in \figref{fig:res_ablation}.

\textbf{Varying the value of $\nu$:} We use the facility location variant of \scg\ (\textsc{Flcg}) to study the effect of $\nu$ for active data discovery and present the results in \figref{fig:res_ablation}(c,d). We observe that higher values of $\nu$ enforce stricter conditioning and leads to finding more data points from the unknown classes, while very small values of $\nu$ may not provide enough conditioning and select lesser samples from the unknown classes. Note that using a value of $\nu$ that is too high may result in selecting data points from a particular unknown class. Intuitively, this class may be the most distant from other classes in the feature space. 

\textbf{Varying the value of $\eta$:} We use the facility location variant of \smi\ (\textsc{Flmi}) to study the effect of $\eta$ for active data discovery and present the results in \figref{fig:res_ablation}(a,b). We observe that higher values of $\eta$ do not necessarily result in stronger targeting and finding more data points from the unknown classes. Infact, values of $\eta$ that are too high may result in lesser data points being select from the unknown classes. Usually, for most datasets, simply setting $\eta=1$ suffices for the targeting phase in active data discovery.



\end{document}